\documentclass[sigconf]{acmart}
\makeatletter
\def\@ACM@checkaffil{
    \if@ACM@instpresent\else
    \ClassWarningNoLine{\@classname}{No institution present for an affiliation}%
    \fi
    \if@ACM@citypresent\else
    \ClassWarningNoLine{\@classname}{No city present for an affiliation}%
    \fi
    \if@ACM@countrypresent\else
        \ClassWarningNoLine{\@classname}{No country present for an affiliation}%
    \fi
}
\makeatother
\usepackage[utf8]{inputenc} 
\usepackage[T1]{fontenc}    
\usepackage{hyperref}       
\usepackage{url}            
\usepackage{booktabs}       
\usepackage{amsfonts}       
\usepackage{nicefrac}       
\usepackage{microtype}      
\usepackage{xcolor}         
\usepackage{soul}

\usepackage{graphicx}
\usepackage{subfig}
\usepackage{booktabs} 
\usepackage{amsthm}
\usepackage{wrapfig}
\usepackage{tabularx}
\usepackage[mathscr]{euscript}
\usepackage{multicol}
\usepackage{multirow}
\usepackage{hhline}
\usepackage{array}
\usepackage{framed}
\usepackage{enumitem}

\theoremstyle{plain}
\newtheorem{theorem}{Theorem}  
\newtheorem{lemma}[theorem]{Lemma} 
\newtheorem{remark}{Remark} 

\theoremstyle{definition}
\newtheorem{assumption}{Assumption} 

\theoremstyle{remark}

\usepackage{algorithm,algpseudocode}
\usepackage{amsmath}
\usepackage{bm}
\usepackage{color}
\usepackage{apxproof}
\usepackage{gensymb}
\usepackage{pifont}

\DeclareMathOperator*{\argmin}{argmin}
\DeclareMathOperator*{\argsort}{argsort}
\usepackage{xspace}
\usepackage{mathtools}

\makeatletter
\DeclareRobustCommand\onedot{\futurelet\@let@token\@onedot}
\def\@onedot{\ifx\@let@token.\else.\null\fi\xspace}

\def\eg{\emph{e.g}\onedot} 
\def\ie{\emph{i.e}\onedot}

\makeatother

\AtBeginDocument{%
  \providecommand\BibTeX{{%
    \normalfont B\kern-0.5em{\scshape i\kern-0.25em b}\kern-0.8em\TeX}}}

\setcopyright{acmcopyright}
\copyrightyear{2018}
\acmYear{2018}
\acmDOI{XXXXXXX.XXXXXXX}

\acmConference[Conference acronym 'XX]{Make sure to enter the correct
  conference title from your rights confirmation emai}{June 03--05,
  2018}{Woodstock, NY}
\acmPrice{15.00}
\acmISBN{978-1-4503-XXXX-X/18/06}

\begin{document}

\title{Everything Perturbed All at Once: \\
Enabling Differentiable Graph Attacks}

\author{Haoran Liu, Bokun Wang, Jianling Wang, Xiangjue Dong, Tianbao Yang, James Caverlee}


\affiliation{%
  \institution{Department of Computer Science and Engineering, Texas A\&M University}
}
\email{liuhr99, bokun-wang, jlwang, xj.dong, tianbao-yang, caverlee@tamu.edu}



\renewcommand{\shortauthors}{Haoran Liu, et al.}

\begin{abstract}

As powerful tools for representation learning on graphs, graph neural networks (GNNs) have played an important role in applications including social networks, recommendation systems, and online web services.
However, 
GNNs have been shown to be vulnerable to adversarial attacks, which can significantly degrade their effectiveness. 
Recent state-of-the-art approaches in adversarial attacks rely on gradient-based meta-learning to selectively perturb a single edge with the highest attack score until they reach the budget constraint. While effective in identifying vulnerable links, these methods are plagued by high computational costs. By leveraging continuous relaxation and parameterization of the graph structure, we propose a novel attack method called Differentiable Graph Attack (DGA) to efficiently generate effective attacks and meanwhile eliminate the need for costly retraining. Compared to the state-of-the-art, DGA achieves nearly equivalent attack performance with 6 times less training time and 11 times smaller GPU memory footprint on different benchmark datasets. Additionally, we provide extensive experimental analyses of the transferability of the DGA among different graph models, as well as its robustness against widely-used defense mechanisms. 
\end{abstract}





\received{20 February 2007}
\received[revised]{12 March 2009}
\received[accepted]{5 June 2009}


\maketitle

\section{Introduction}

Graph Neural Networks (GNNs)~\cite{hamilton2017inductive, kipf2017semisupervised} are powerful in modeling graph-structured data and show remarkable performance on many real-world applications such as
social networks~\cite{fan2019graph, rozemberczki2021multi, tang2022friend}, recommendation systems~\cite{chen2022graph, pang2022heterogeneous, wang2021graph, ying2018graph, zhao2022joint}, 
and drug discovery~\cite{ingraham2019generative, stark2022equibind}.
Given the successful applications of GNNs, there are also growing concerns about their robustness under adversarial attacks~\cite{bojchevski2019adversarial, dai2018adversarial, ma2022adversarial, zugner2019adversarial, wang2021certified}.
For example, toxic behavior detectors with GNN backbones could be vulnerable to adversarial attacks, leading to undetected instances of harassment, extremism, or radicalization targeting innocent individuals on social media~\cite{deng2021graph, hu2022detecting}.
To ensure the reliability and safety of GNN-based systems in practice, it becomes paramount to understand their vulnerabilities to adversarial attacks as a foundation for their robust deployment.
In a nutshell, developing more effective adversarial attack methods on GNNs not only aids in assessing the robustness and defense strategies of GNNs~\cite{geisler2021robustness, jin2020graph, tang2020transferring, zhang2022robust, zhang2020gnnguard, zhu2021shift} but also enhances the understanding of the underlying properties of current GNN models. 

Presently, adversarial attacks on GNNs can be categorized based on the attacker's capacity, goal, level of knowledge, and perturbation type~\cite{jin2021adversarial, sun2022adversarial, xu2020adversarial}. 
One of the most practical setups among these is the \textbf{gray-box attack}, in which attackers have complete knowledge of the training data but no information on the victim model. Following this setup as in previous works ~\cite{liu2022gradients, liu2022towards, zugner2019adversarial}, within the budget limit, we seek for attacking strategies which perform topology attacks (\ie adding or removing edges) on poisoning training data to compromise the overall node classification performance of the victim model.


Despite the effectiveness in identifying vulnerable edges via the gradient-based \textit{``learn to attack''} approaches, the state-of-the-art  methods~\cite{liu2022towards, zugner2019adversarial} in graph adversarial attacks are faced with two major bottlenecks: (i) \textbf{Non-convexity and discrete structure}:  
Meta-learning approaches usually formulate the attack as a bi-level optimization problem, which is challenging to solve due to the non-convexity of both levels, making it difficult to derive a closed-form solution. 
Additionally, the discrete nature of the graph structure poses another challenge for directly applying widely-used gradient-based techniques such as FGSM~\cite{goodfellow2015explaining} and PGD~\cite{madry2018towards}, which are typically employed on image data; and 
(ii) \textbf{Computational and resource costs}: Existing methods selectively perturb a single edge with the highest attack score and require retraining the surrogate model from scratch, incurring high computational costs. Additionally, it is unrealistic to scale these techniques to handle large datasets. 
\begin{figure}[]
    \centering
    \includegraphics[width=0.4\textwidth]{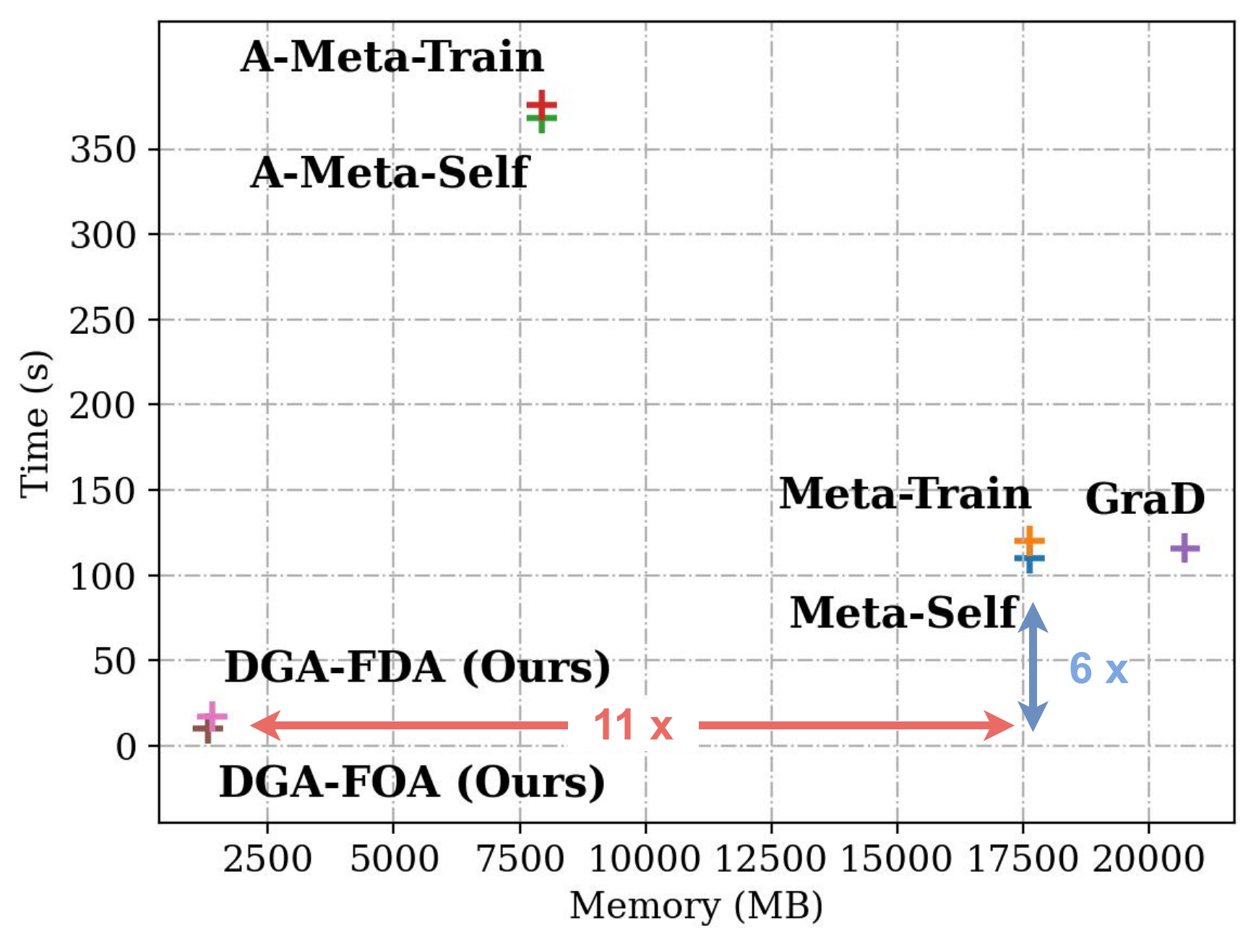}
    \caption{Comparison on computation and resource complexity between MetAttack~\cite{zugner2019adversarial}, GraD~\cite{liu2022towards} and two variants of our DGA method on CiteSeer with 5\% perturbation rate. }
    \label{fig:compare_complex}
    \vspace{-2 pt}
\end{figure}

\begin{figure*}[tp]
    \centering
    \includegraphics[width=1\textwidth]{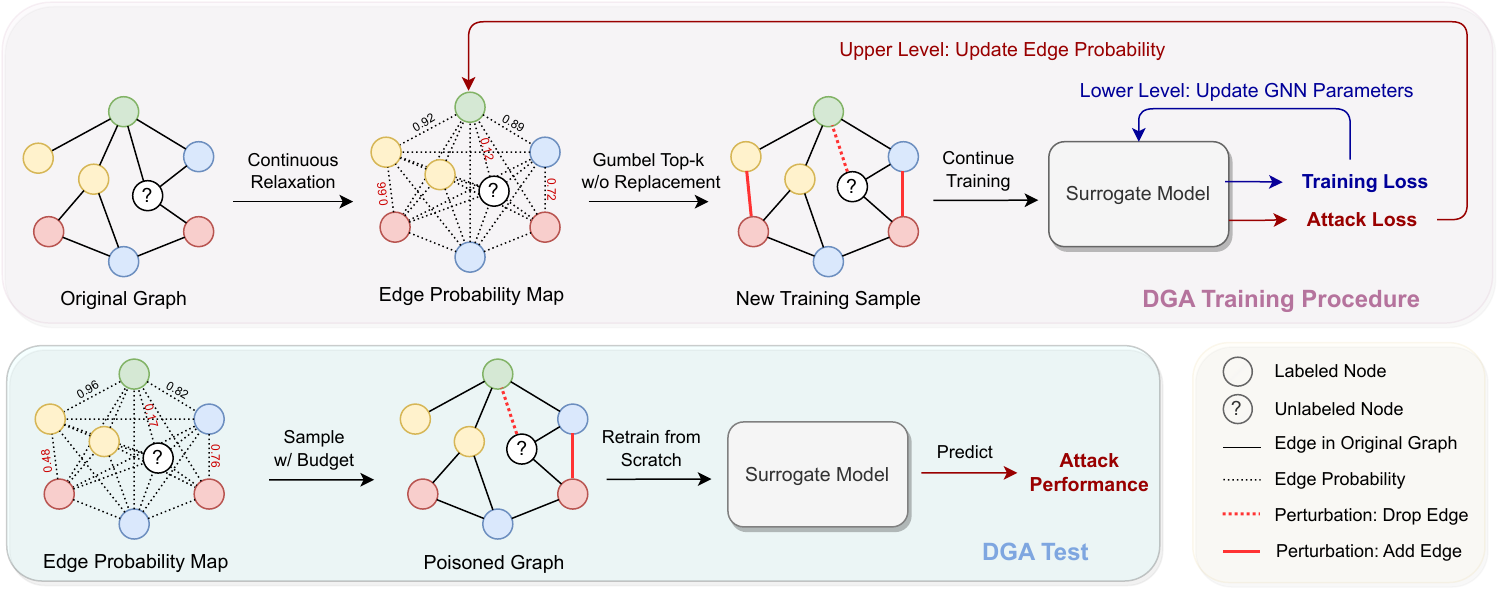}
    \vspace{-10 pt}
    \caption{Illustration of the DGA framework. Node colors indicate different classification labels. }
    \label{fig:dga_framework}
    \vspace{-5 pt}
\end{figure*}


To tackle the above-mentioned bottlenecks, we propose the \underline{\textbf{D}}ifferentiable \underline{\textbf{G}}raph \underline{\textbf{A}}ttack (DGA) with continuous relaxation on the graph structure. It offers several advantages with its novel \textit{train-then-sample} attack scheme. 
During the training process, instead of searching over a discrete set of candidate perturbations, we \textit{relax} the search space to a continuous space, so that the graph structure can be optimized with respect to the attack objective through gradient descent, allowing for fine-grained adjustments and improved effectiveness of the attack. Then, we simulate the lower-level optimization process with one-step fine-tuning on the surrogate model to avoid calculating the accumulation of meta-gradients and further reduce computation complexity. This offers more control over the attack process compared to methods with fixed attack epochs. Essentially, DGA utilizes a single-step adaption approach to simulate the poisoning process, offering a trade-off between the poisoning simulation procedure and training time. For the sampling stage, our sampling scheme enables DGA to generate poisoned graphs with varying budgets while being trained only once, further streamlining the attack process. 
Theoretically, we show that the estimation error in forward/backward passes of our algorithm, as well as the convergence of the method, depends on the error of edge sampling. These benefits collectively make DGA a powerful and efficient method for generating adversarial attacks on graph structures, as shown in Figure~\ref{fig:compare_complex}. 
``FOA'' and ``FDA'' refer to first-order approximation and finite-difference approximation, respectively (See Section~\ref{sec:cnts_relax} for details). 
We conduct empirical analysis of DGA with common benchmark datasets. Our main contributions are summarized as follows: 

\begin{itemize}
[topsep=5pt,parsep=5pt,partopsep=5pt, leftmargin=8mm]
    \item \textbf{Effectiveness.} DGA outperforms or performs comparably to the SOTA methods on widely-used datasets. Notably, our method is effective even when the perturbation budget is low, which is more aligned with real-world scenarios and maintains the unnoticeability of the attack.
    \item \textbf{Time and Resource Efficiency.} DGA is more than 6 times faster and requires approximately 11 times less GPU memory than SOTA methods.  Importantly, the attack time of DGA remains constant even as the perturbation budget increases. Additionally, DGA can be easily adapted to large-scale graphs, making it a practical choice for real-world scenarios.
    \item \textbf{Transferability and Imperceptibility.}  DGA exhibits excellent transferability to a variety of GNN models and demonstrates imperceptibility toward commonly-used defense methods. These findings highlight the valuable role of DGA in assessing the robustness of existing GNN methods and providing insights for the development of novel defense strategies against adversarial attacks.
\end{itemize}





\section{Background} \label{sec:hierarchical_rep}


 We consider the task of semi-supervised node classification, in which a model is trained to predict the labels of nodes in a graph, using both labeled and unlabeled data. 
 We denote an undirected graph as $G=\left(\mathcal{V}, \mathcal{E}\right)$, 
 where $\mathcal{V} = \{i= 1, \dots, N\}$ is the set of $N$ nodes,  and $\mathcal{E} \subseteq \mathcal{V} \times \mathcal{V}$ is the edge set with cardinality $|\mathcal{E}| = M$. 
 In the semi-supervised setting, we are given a subset $\mathcal{V}_L\subseteq \mathcal{V}$ of labeled nodes, in which the nodes are associated with class labels from $C=\{c_1, c_2, \cdots, c_k\}$. The goal is to learn a function $f$ to infer labels of nodes in the unlabeled node set $\mathcal{V}_U=\mathcal{V}\setminus \mathcal{V}_L$.
 Without loss of generality, we collectively use $X \in \mathbb{R}^{N\times D}$ to denote the node features of all nodes in the graph, where $D$ is the dimension of feature vectors. 
 Additionally, we denote the adjacency matrix associated with this graph as $A \in \{0, 1\}^{N\times N}$. For each entry $(i, j)$ in $A$, $A_{i,j}=1$ if $(i,j)\in \mathcal{E}$ and $0$ otherwise.
 In this context, the neighborhood of node $v$ is denoted as $\mathcal{N}(v) = \{i\in \mathcal{V}: (i, v)\in \mathcal{E}\}$. 
 Given a graph $G$, the node embeddings are obtained by using a GNN model $f_{\theta}(X, A)$, parameterized by $\theta$. Particularly, the GNN model takes the node features $X$ and adjacency information $A$ as input and outputs the logits of each node. 


\subsection{Problem Setting} \label{sec:probset}

We follow \cite{zugner2019adversarial} and focus on the \textit{combination} of the following specific attack setting:
(1) \textbf{gray-box}, in which the attackers have access to complete information about the training data but zero knowledge about the specifics of the underlying model. To overcome this challenge, surrogate models are utilized to approximate and simulate the behavior of the target model;
(2) \textbf{poisoning}, in which the attacker's goal is to increase the classification error (\ie one minus accuracy) by training on modified (\ie poisoned) data;
(3) \textbf{graph structure attack}, in which the attacker's perturbation type is adding/removing edges.
To ensure the attack remains undetected, a maximum perturbation budget denoted as $\Delta$, which restricts the difference between the perturbed graph structure $A$ and the original structure $A_{orig}$ such that $\|A-A_{orig}\|_0 \leq 2\Delta$;
(4) \textbf{untargeted/global}, in which the attacker's goal is to compromise the \textit{overall} node classification performance of the model instead of targeting individual nodes.

Formally, this specific graph attack setting can be formulated as a \textit{bi-level optimization} problem
\begin{equation}
    \begin{aligned}
& \min_{A\in \mathcal{A}_{\Delta}} \ell_{atk}\left(f_{\theta^*}(X, A), Y\right), \\
& \textrm{s.t.} \quad \theta^* = \arg \min_\theta \ell_{train}(f_\theta(X, A), Y),
    \end{aligned} 
\label{eq:bi-level}
\end{equation}
where $\mathcal{A}_{\Delta}\coloneqq \{A' \in \{0,1\}^{N\times N}\mid \|A'-A_{orig}\|_0 \leq 2\Delta\}$ denotes the search space containing all possible adjacency matrices with the given modification budget $\Delta$. 
Here, the upper-level optimization aims to find the optimal edge perturbations which result in a new adjacency matrix $A$ that maximizes the attack success. The attack objective $\ell_{atk}$ could be the inverse of either training loss $\ell_{train}$ or self-training loss $\ell_{self}$. Specifically, $\ell_{train}$ is the cross-entropy loss on the training set and $\ell_{self}$ is computed with pseudo labels for the unlabeled nodes. The pseudo labels are predicted by a well-trained surrogate model given the clean graph. In a sense, $\ell_{self}$ can be used to estimate the generalization loss after the attack.  
As for the nested inner problem, the lower-level optimization targets finding the optimal GNN parameters $\theta^*$ through an optimization process of training the GNN on the perturbed graph from scratch. Essentially, finding a good attack method for GNNs involves solving this bi-level optimization problem in a effective and efficient manner.

\section{Differentiable Graph Attack}\label{sec:dga}
In light of the non-convex and combinatorial characteristics of the optimization problem presented in Equation \eqref{eq:bi-level}, it is infeasible to acquire a closed-form solution or employ gradient-based iterative algorithms commonly used in deep learning.
To address this challenge, we propose a novel approach that is both effective and efficient by 
continuous relaxing and parameterizing the graph structure as a continuous edge probability map. 
An illustration of the DGA framework is shown in Figure~\ref{fig:dga_framework}.

\subsection{Continuous Relaxation on Graph Structure}\label{sec:cnts_relax}

We model each edge with a Bernoulli random variable and transform the discrete graph adjacency matrix $A$ to an unnormalized probability matrix $P$, where the $(i, j)$-th element $p_{ij}$ indicates the (unnormalized) edge probability between node $i$ and node $j$ under the attack setting. Then, the search space for modified adjacency matrix $\{0,1 \}^{N\times N}$ is relaxed to the positive orthant $\mathbb{R}_{++}^{N\times N}$. 
As shown in Figure~\ref{fig:dga_framework}, there is no need to enforce the perturbation budget constraint during the DGA training process. The technique outlined in Section~\ref{sec:sample} ensures that the generated poisoned graph consistently adheres to the budget. We consider the following continuous relaxation of \eqref{eq:bi-level}.
\begin{equation}
\begin{aligned}
    \label{eq:bi-level-cnts}\min_{P\in\mathbb{R}_{++}^{N\times N}} \ell_{atk}(f_{\theta^*}(X,P),Y),\quad \quad \\
    \mathrm{s.t.} \quad \theta^* = \argmin_{\theta} \ell_{train}(f_\theta(X,P),Y).	
\end{aligned}
\end{equation}
Instead of exactly solving the lower-level problem,  we approximate the optimal surrogate model $\theta^*$ by a single-step adaptation with step size $\alpha$ following the techniques in~\citet{liu2018darts,finn2017model}. 
By changing variables $Q = \log P$, the optimization problem in \eqref{eq:bi-level-cnts} becomes unconstrained w.r.t. unnormalized log-probabilities $Q$. We define that $\mathbf{q}\coloneqq \mathrm{vec} (Q)$, $\mathcal{L}_{atk} (\theta,\mathbf{q}) \coloneqq \ell_{atk}(f_{\theta}(X,P),Y)$, $\mathcal{L}_{train} (\theta,\mathbf{q}) \coloneqq  \ell_{train}(f_{\theta}(X,P),Y)$, and $\nabla_1 \mathcal{L}(\cdot,\cdot)$, $\nabla_2 \mathcal{L}(\cdot,\cdot)$ are gradients of $\mathcal{L}$ with respect to its first and second arguments, respectively. Then, the bi-level optimization in \eqref{eq:bi-level-cnts} becomes a simpler compositional optimization problem.
\begin{equation}\label{eq:composition}
\min_{\mathbf{q}\in\mathbb{R}^{N\times N}} \mathcal{L}_{atk}(\hat{\theta},\mathbf{q}),\quad \hat{\theta} = \theta - \alpha \nabla_{\theta} \mathcal{L}_{train}(\theta, \mathbf{q}).
\end{equation}
The hyper-gradient with respect to the log probabilities $\mathbf{q}$ can be computed as
\begin{align}\label{eq:hyper-grad}
\nabla_{\mathbf{q}} \mathcal{L}_{atk}(\hat{\theta},\mathbf{q}) = \nabla_2\mathcal{L}_{atk}(\hat{\theta},\mathbf{q}) - \alpha \nabla^2_{1,2}\mathcal{L}_{train}(\theta,\mathbf{q}) \nabla_1 \mathcal{L}_{atk}(\hat{\theta},\mathbf{q}).
\end{align}
Based on the hyper-gradient, 
we can update the log-probability matrix $Q$ by a gradient descent step. 
It is worth noting that the Hessian matrix computation in \eqref{eq:hyper-grad} can be costly when the number of nodes is large. To this end, we can either set $\alpha = 0$ in the hyper-gradient (``first-order approximation'') or use the finite difference approximation similar to \citet{liu2018darts}. To be specific, with $\theta_{\pm}= \theta \pm \delta\nabla_1\mathcal{L}_{atk}(\hat{\theta},\mathbf{q})$, we can approximate the last term in \eqref{eq:hyper-grad} as
\begin{align*}
\nabla^2_{1,2}\mathcal{L}_{train}(\theta,\mathbf{q}) \nabla_1 \mathcal{L}_{atk}(\hat{\theta},\mathbf{q}) \approx \frac{\nabla_2 \mathcal{L}_{train}(\theta_+,\mathbf{q})-\nabla_2 \mathcal{L}_{train}(\theta_-,\mathbf{q})}{2\delta}.
\end{align*}
In the rest of the paper, we refer to the variant of our algorithm using first-order approximation ($\alpha = 0$ in Eq.~\ref{eq:hyper-grad}) as \textbf{DGA-FOA}, and the variant using the finite difference approximation as\textbf{ DGA-FDA}.

\algnewcommand{\Initialize}[1]{%
	\State \textbf{Initialize:}
	\Statex \hspace*{\algorithmicindent}\parbox[t]{.8\linewidth}{\raggedright #1}
}

\begin{algorithm}[tbp]
	\caption{Training Procedure of Differentiable Graph Attack (DGA)}\label{alg:DGA-train} 
	\begin{algorithmic}[1]
		\State\textbf{Input:} Graph structure $A_{orig}$, step size $\eta$, total num. of iterations $T$, node features $X$, labels $Y$
		\Initialize{$Q_0 \leftarrow \log (A_{orig} + \epsilon)$\quad~     \Comment{\small  Add an $\epsilon$ to avoid numerical issue}\\
			$\theta \leftarrow \argmin_\theta \ell_{train}(f_\theta(X, A_{orig}), Y)$}
		\For {$t=0,1,\dotsc, T-1$}
		\State Sample a sparse graph $\tilde{A}_t$ from unnormalized log-probabilities $Q_t$ by Gumbel top-$k$ sampling 
	  \State Compute the single-step adaptation by  $\hat{\theta}_t  = \theta -\alpha \nabla_{\theta}\ell_{train}(f_{\theta}(X, \tilde{A}_t), Y )$ 
	 \State Compute the approximated hyper-gradient $\tilde{\nabla}_t$ based on $\hat{\theta}_t$ and $\tilde{A}_t$: Option I: first-order approximation (FOA) or Option II: finite-difference approximation (FDA)
\State Update the probability matrix by $Q_{t+1} = Q_t - \eta \tilde{\nabla}_t$ and symmetrize $Q_{t+1}$
		\EndFor
		\State Obtain the probability matrix $P$ from the unnormalized log-probability matrix $Q_T$
		 \State\textbf{Return:} Probability matrix $P$ for graph poisoning
	\end{algorithmic}
\end{algorithm}

\subsection{Edge Sampling for Expressiveness and Efficiency}\label{sectin:edge_sampling}

Note that the gradient-based continuous optimization of \eqref{eq:composition} results in a dense probability matrix $P$. Utilizing this matrix as input may lead to over-smoothing~\cite{hamilton2017inductive, Zhao2020PairNorm:} or over-squashing~\cite{di2023over} of the surrogate graph neural network $\theta$.  Moreover, message passing on a dense graph exacerbates the issue of ``neighborhood explosion'', which significantly increases computational costs. To resolve this problem, we use the Gumbel-Top-$k$ trick~\citep{kool2019stochastic,xie2019reparameterizable} to sample a sparser graph $\tilde{A}$ from the unnormalized log-probabilities $Q$. Formally, for the $i$-th node $v_i$, we construct $k$ edges between $v_i$ and each of the first $k$ elements of
\begin{equation*}
	\begin{aligned}
		\argsort_j \left(- \log(- \log(u_j))+Q_{ij}\right),
	\end{aligned} 
\end{equation*}
where $u_j$ is a vector randomly sampled from the uniform distribution. The top-$k$ operation can be replaced by the differentiable relaxed top-$k$ operation in \citet{xie2019reparameterizable}. Intuitively, such sampling can be regarded as a stochastic relaxation of the $k$-nearest neighbors rule.

\subsection{Obtaining the Poisoned Graph} \label{sec:sample}
Utilizing the learned edge probability map $P$ (output of Alg.~\ref{alg:DGA-train}), we generate perturbations by sampling from the discrepancy between the final and initial edge probability maps. This process serves to test the performance of the DGA attack, as illustrated in the blue box in Figure~\ref{fig:dga_framework}.
To prepare for the attack, DGA first symmetrizes the edge probability matrix $P$ for undirected graphs as $\bar{P} = \frac{1}{2}(P + P^\top)$. Next, we compute the difference score matrix $S = (\bar{P} - A_{orig})\odot (1 - 2 A_{orig})$, where $\odot$ indicates element-wise multiplication. 
Here we flip the sign of difference score for every connected edge in the original graph $A_{orig}$, which allows us to identify existing edges with the most significant negative impact, as well as the non-existent edges with the most potent positive influence on the attack objective. 
The resulting difference score matrix $S$ is employed to construct a categorical distribution, from which we can sample perturbations in order to obtain the poisoned graph.


Taking into account the allocated attack budget, we sample $\Delta$ edges from the categorical distribution without replacement and flip them, thereby generating a poisoned graph. Subsequently, we proceed to train models from scratch using this poisoned graph, allowing us to assess the attack's performance on a specific model. In practice, we repeat this sampling process multiple times and select the instance that yields the best $\ell_{atk}$ (attack loss) value.

\section{Convergence Analysis}

Before the $t$-th iteration, our algorithm samples edges for each node $i$ by the Gumbel top-$k$ trick, which is equivalent to sampling $k$ edges from $\mathbf{p}_t^i = \left(p_t^{i,1}, \dotsc, p_t^{i,N}\right)$ without replacement~\citep{kool2019stochastic}, where $\mathbf{p}_t^i$ is the normalized probabilities corresponding to the $i$-th line of $Q_t$. This result in the sparsified log-probability matrix $\tilde{Q}_t  = \textrm{Diag}(\tilde{\mathbf{q}}_t)$ and $\tilde{\mathbf{q}} \in \{0,1\}^{N^2}$ is a vector that stores sampled edges for each node, i.e. $\tilde{\mathbf{q}}_t^{i\cdot j} =\mathbf{q}_t^{i\cdot j}$ if the edge $(i,j)$ is sampled and $\tilde{\mathbf{q}}_t^{i\cdot j} = 0$ otherwise. Note that the sampled adjacency matrix $\tilde{A}_t$ is just the binarized copy of $\tilde{Q}_t$. For brevity, we define that $\Phi(\mathbf{q})\coloneqq \mathcal{L}_{atk}(\hat{\theta},\mathbf{q})$, $\hat{\theta}= \theta - \alpha \nabla_\theta \mathcal{L}_{train}(\theta,\mathbf{q})$. We make some regularity assumptions of the loss functions.
\begin{assumption}\label{asm:lip_short}
Suppose that the loss function $\mathcal{L}_{atk}$ is Lipschitz continuous and has Lipschitz-continuous gradient while $\mathcal{L}_{atk}$ has Lipschitz continuous gradient and Hessian. 
\end{assumption}

\begin{lemma}\label{lem:smooth}
Under Assumption~\ref{asm:lip_short}, there exists $G>0$ and $L>0$ such that $\Phi$ is $G-$Lipschitz continuous and $\nabla \Phi$ is $L$-Lipschitz continuous.
\end{lemma}
The proof of Lemma~\ref{lem:smooth} can be found in Appendix~\ref{appendix:proofs}.
With the sparsified $\tilde{\mathbf{q}}$, we approximate the objective function $\Phi(\mathbf{q})$ by $\Phi(\mathbf{\tilde{q}})$ and the error is
\begin{align}\label{eq:est_err}
\|\Phi(\mathbf{\tilde{q}_t}) - \Phi(\mathbf{q}_t)\|_2 \leq G\|\mathbf{\tilde{q}}_t - \mathbf{q}_t\|_2,
\end{align}
where $\|\mathbf{\tilde{q}}_t - \mathbf{q}_t\|_2$ is the estimation error by cause of the edge sampling per iteration. Besides, the hyper-gradient $\nabla \Phi(\mathbf{q})$ in \eqref{eq:hyper-grad} can be estimated by 
\begin{align*}
\nabla  \Phi(\mathbf{\tilde{q}}) &= \nabla_2 \mathcal{L}_{atk}(\hat{\tilde{\theta}},\mathbf{\tilde{q}}) - \alpha \nabla^2_{1,2}\mathcal{L}_{train} (\theta,\mathbf{\tilde{q}})\nabla_1 \mathcal{L}_{atk}(\hat{\theta},\mathbf{\tilde{q}}),\quad \\
\hat{\tilde{\theta}} &= \theta - \alpha \nabla_\theta \mathcal{L}_{train}(\theta,\mathbf{\tilde{q}}),
\end{align*}
where $\nabla  \Phi(\mathbf{\tilde{q}})$ is the vectorized $\tilde{\nabla}_t$ in Alg.~\ref{alg:DGA-train}. Next, we present the main theorem on convergence.

\begin{theorem}\label{thm:convergence}
Under assumptions above, our algorithm with proper step size $\eta$ leads to
\begin{align}\label{eq:bound}
\mathbb{E}\|\nabla \Phi(\mathbf{q}_\tau)\|_2^2 \leq \mathcal{O}\left(\frac{\Phi(\mathbf{q}_0) - \inf_{\mathbf{q}}\Phi}{T} + \mathrm{Avg. Err.}\right),
\end{align}
where $\tau$ is sampled from $\{0,1,\dotsc, T-1\}$ uniformly at random and $\mathrm{Avg. Err.}$ is the average error due to edge sampling across the iterations. To be specific, $\mathrm{Avg. Err.}\coloneqq \frac{1}{T}\sum_{t=0}^{T-1}\mathbb{E} \|\tilde{\mathbf{q}}_t - \mathbf{q}_t\|_2^2$.
\end{theorem}
\begin{remark}
The proof of Theorem~\ref{thm:convergence} can be found in Appendix~\ref{appendix:proofs}. Theorem~\ref{thm:convergence} demonstrates that our algorithm exhibits an $\mathcal{O}(1/T)$ non-asymptotic convergence rate when $k=N$, the same as that of gradient descent. While opting for a larger value of $k$ can be advantageous from the estimation/optimization perspective, it results in worse expressiveness and efficiency as explained in Section~\ref{sectin:edge_sampling}. Therefore, we can manage the trade-off between estimation/optimization error and expressiveness/efficiency by adjusting the hyper-parameter $k$. Fortunately, the estimation error $\|\mathbf{\tilde{q}}_t - \mathbf{q}_t\|_2$ could be negligible even with a small $k$ when the distribution constructed by $\mathbf{p}_i$ is light-tailed (e.g., power law), thanks to the Gumbel top-$k$ sampling.
\end{remark}

\begin{table}[tbp]\centering
\vspace{-5pt}
\caption{Dataset statistics.}\label{tab:datastat}
\vspace{-5pt}
\resizebox{0.49\textwidth}{!}{
\begin{tabular}{lccccc}\toprule
\textbf{Dataset} &\textbf{\#Nodes} &\textbf{\#Edges} &\textbf{\#Classes} &\textbf{\#Features} &\textbf{\# Avg. Deg.} \\\midrule
\textbf{Citeseer} &2,110 &3,757 &6 &3,703 &1.43 \\
\textbf{Cora} &2,485 &5,069 &7 &1,433 &2.00 \\
\textbf{PolBlogs} &1,222 &16,714 &2 &-- &13.68 \\
\bottomrule
\end{tabular}}
\vspace{-5pt}
\end{table}

\begin{table*}[tbp]
\centering 
\vspace{-5pt}
\caption{Test accuracy (\%) of the \textit{GCN} model after training with clean and poisoned graphs. The average performance are calculated based on 10 runs. 
The top-two results are highlighted as \textbf{1st} (bold) and \underline{2nd} (underlined). The standard deviations are provided in Appendix~\ref{appendix: experiment_result}.
} \label{tab:main-res}
\resizebox{0.9\textwidth}{!}{
\begin{tabular}{lcccc|cccc|cccc}  
\toprule
\textbf{Dataset}                & \multicolumn{4}{c|}{\textbf{CiteSeer}}                                  & \multicolumn{4}{c|}{\textbf{Cora}}                             & \multicolumn{4}{c}{\textbf{PolBlogs}}                         \\
\textbf{Perturbation Rate (\%)} & \textbf{0} & \textbf{1}     & \textbf{3}     & \textbf{5}     & \textbf{0} & \textbf{1}     & \textbf{3}     & \textbf{5}     & \textbf{0} & \textbf{1}     & \textbf{3}     & \textbf{5}     \\ 
\midrule
DICE~\cite{waniek2018hiding}                   & 71.81      & 71.40           & 70.73          & 70.05          & 83.62      & 82.43          & 81.96          & 81.45          & 95.00         & 92.41          & 89.29          & 86.85          \\
Meta-Self~\cite{zugner2019adversarial}                & 71.81      & 71.48          & 69.3           & \textbf{66.91} & 83.62      & 82.08          & \underline{78.88}          & \textbf{75.46} & 95.00         & \textbf{85.02} & \textbf{79.17} & \textbf{76.40}  \\
Meta-Train~\cite{zugner2019adversarial}                & 71.81      & 70.50           & 69.31          & 67.86          & 83.62      & 82.16          & 79.99          & 77.69          & 95.00         & 92.42          & 88.85          & 88.46          \\
A-Meta-Self~\cite{zugner2019adversarial}              & 71.81      & 71.29          & 70.83          & 69.97          & 83.62      & 83.32          & 82.45          & 81.69          & 95.00         & 94.30           & 92.90           & 92.16          \\
A-Meta-Train~\cite{zugner2019adversarial}              & 71.81      & 70.60           & 69.02          & 67.54          & 83.62      & 82.18          & 79.24          & 77.12          & 95.00         & 94.39          & 92.10           & 90.45          \\
GraD~\cite{liu2022towards}              & 71.81      & 71.64 & 70.88 & 70.66          & 83.62      & 83.43	& 82.77	& 82.29          & 95.00         & 91.77 &	89.01 & 88.71          \\
\midrule
DGA-FOA (ours)       & 71.81      & \underline{68.75} & \textbf{67.89} & \textbf{66.91} & 83.62      & \textbf{79.89} & \textbf{78.77} & \underline{78.07}          & 95.00         & 91.31          & \underline{88.45}          & \underline{87.01}          \\
DGA-FDA (ours)      & 71.81      & \textbf{68.51} & \underline{68.15} & \underline{67.45} & 83.62      & \underline{81.16}          & 80.32          & 79.50           & 95.00         & \underline{90.90}           & 89.26          & 87.63     \\   \bottomrule 
\end{tabular}}
\end{table*}

\begin{table}[bp]\centering
\caption{Comparison of training time (in seconds), GPU memory occupancy (in MB) after attack between DGA and baselines with 5\% perturbation rate. All experiments are conducted on a single 24GB NVIDIA RTX A5000 GPU for a fair comparison. }\label{tab: computation_complexity}
\resizebox{0.5\textwidth}{!}{
\begin{tabular}{lcc|cc|cc}\toprule
\textbf{Dataset} &\multicolumn{2}{c|}{\textbf{Citeseer}} &\multicolumn{2}{c|}{\textbf{Cora}} &\multicolumn{2}{c}{\textbf{PolBlogs}} \\
\textbf{Method} &\textbf{Time} &\textbf{Mem.} &\textbf{Time} &\textbf{Mem.} &\textbf{Time} &\textbf{Mem.} \\
\midrule
Meta-Self~\cite{zugner2019adversarial} &114&17,629  &172 &21,993 &225 &18,453 \\
Meta-Train~\cite{zugner2019adversarial} &114&17,629  &172 &21,993 &225 &18,453 \\
A-Meta-Self~\cite{zugner2019adversarial} &368&7,933  &692 &13,375 &462 &12,289 \\
A-Meta-Train~\cite{zugner2019adversarial} &376&7,933  &692 &13,375 &462 &12,289 \\
GraD~\cite{liu2022towards} &116  &20,723   &176   &23,515   &138   &19,163 \\
\midrule
DGA-FOA (ours)&\textbf{14} &\textbf{1,327} &\textbf{15} &\textbf{1,563} &13 &\textbf{1,109} \\
DGA-FDA (ours) &17 &1,365 &24 &1,635 &\textbf{8} &1,129 \\
\bottomrule
\end{tabular}}
\end{table}

\section{Experiments}\label{sec: exp}
Our proposed attack method, DGA (Differentiable Graph Attack), is evaluated through a series of experiments aimed at demonstrating its effectiveness and efficiency. 
The experimental settings are introduced in Section~\ref{sec:exp_setting}. The attack performance and the generalizability and transferability of DGA is analyzed in Section~\ref{sec:attack}. 
We also show the computation complexity of DGA in Section~\ref{sec:complex}. 
Furthermore, we conduct experiments with defense methods in Section~\ref{sec:defense} to demonstrate the robustness of our approach. 
Additionally, we present a study on Gumbel-$k$ in Section~\ref{sec:gumbel-k} and provide visualization of the distribution of the generated poisoned graph in Section~\ref{sec:visual}.


\subsection{Experimental Settings} \label{sec:exp_setting}

\textbf{Dataset.} The experiments are conducted on three widely-used datasets: two citation network datasets, namely CiteSeer~\cite{sen2008collective} and Cora~\cite{mccallum2000automating}, and one social network dataset PolBlogs~\cite{adamic2005political}. All experiments are performed on the largest connected component of the graphs.
Following previous works~\cite{jin2020graph, zugner2018adversarial}, we randomly split the datasets into the train, validation, and test sets using a 10\%, 10\%, and 80\% ratio, respectively. 
For each experiment, we report the average performance of 10 runs. 
The experiments are conducted on the largest connected component (LCC) of the graphs, ensuring that only the main connected portion of the graphs is considered for analysis and evaluation. The statistics for the LCC of these datasets are summarized in Table~\ref{tab:datastat}. 

\vspace{3pt} \noindent 
\textbf{Victim Models.} 
Following previous methods, we first take the widely-used GCN~\cite{kipf2017semisupervised} as our victim model. 
Moreover, as we adopt a gray-box setting, in which the model architecture is considered unknown, we further conduct experiments with Graph
Attention Network (GAT)~\cite{veličković2018graph} and DeepWalk~\cite{perozzi2014deepwalk} to measure the transferability of DGA. 
Additionally, to assess the effectiveness of our proposed DGA attack method, we conduct experiments to evaluate its robustness against existing defense methods, namely GCN-Jaccard and GCN-SVD.

\vspace{3pt} \noindent 
\textbf{Perturbation Budgets.} To be closer to real-world scenarios and demonstrate the imperceptibility of the attack method, we set 1\%, 3\%, and 5\% as the perturbation budget in our experiments. 
Each method is allowed to modify 1\%, 3\%, and 5\% of the number of edges in the original graph, respectively.

\vspace{3pt} \noindent \textbf{Baselines.} 
To assess the effectiveness and applicability of DGA, we conduct a comprehensive comparison with six representative baselines: including heuristic-based method DICE~\cite{waniek2018hiding} and learning-based methods Meta-Self, Meta-Train, A-Meta-Self, A-Meta-Train~\cite{zugner2019adversarial}, and GraD~\cite{liu2022towards}. 


Additional information about the experimental setup, implementation, and baselines can be found in Appendix~\ref{appendix:experiment_setup}. \footnote{The code will be be made publicly available once accepted.} 

\begin{table*}[tbp] 
\caption{An evaluation of DGA in generalizability, transferability, and robustness against existing defense methods. The first two sections show the test accuracy (\%) of \textit{GAT}~\cite{veličković2018graph} and \textit{DeepWalk}~\cite{perozzi2014deepwalk} models trained with both clean and poisoned graphs, where the surrogate model is GCN. The last two sections show the test accuracy (\%) of GCN models trained with clean and poisoned graphs and subsequently vaccinated with two commonly-used defense mechanisms, namely low-rank SVD approximation~\cite{entezari2020all} (GCN-SVD) and Jaccard (GCN-Jaccard)~\cite{xu2019topology}.
To ensure comprehensive evaluation, we conduct 10 runs of experiments and report the average performance. 
The top-two results are highlighted as \textbf{1st} (bold) and \underline{2nd} (underlined). 
The standard deviations are provided in Appendix~\ref{appendix: experiment_result}.
Note that we cannot perform Jaccard on PolBlogs, as this dataset does not contain node features.} \label{tab:more_res}
\resizebox{\textwidth}{!}{
\begin{tabular}{llcccc|cccc|cccc} \toprule
\multirow{2}{*}{\textbf{Victim Model}} 
& \textbf{Dataset}                & \multicolumn{4}{c|}{\textbf{CiteSeer}}                                  & \multicolumn{4}{c|}{\textbf{Cora}}                            & \multicolumn{4}{c}{\textbf{PolBlogs}}                                           \\
& \textbf{Perturbation Rate (\%)} & \textbf{0} & \textbf{1}     & \textbf{3}     & \textbf{5}     & \textbf{0} & \textbf{1}    & \textbf{3}     & \textbf{5}     & \textbf{0} & \textbf{1}           & \textbf{3}           & \textbf{5}           \\ \midrule
\multirow{8}{*}{\textbf{GAT~\cite{veličković2018graph}}} &
DICE~\cite{waniek2018hiding}                   & 73.51      & 73.71          & \textbf{72.67} & 72.18          & 84.06      & 84.09         & 83.45          & 82.92          & 94.97      & 93.81                & 92.12                & \textbf{90.97}                \\
&Meta-Self~\cite{zugner2019adversarial}                & 73.51      & 73.87          & 73.98          & 73.42 & 84.06      & 83.85         & 83.52          & 83.60  & 94.97      & \underline{93.27}       & 92.33       & 91.57       \\
&Meta-Train~\cite{zugner2019adversarial}                & 73.51      & 73.64          & 74.19          & 73.87          & 84.06      & 83.93         & 83.96          & 83.27          & 94.97      & 93.88                & \textbf{91.57}                & 91.81                \\
&A-Meta-Self~\cite{zugner2019adversarial}             & 73.51      & 73.05 & 72.71          & 72.32          & 84.06      & 83.97         & 83.34          & \textbf{82.11} & 94.97      & 93.91                & 92.44                & \underline{91.10}                 \\
&A-Meta-Train~\cite{zugner2019adversarial}             & 73.51      & 73.96          & 73.86          & 73.15          & 84.06      & 84.08         & 83.37          & 82.84          & 94.97      & 94.58                & 93.93                & 93.10                 \\
&GraD~\cite{liu2022towards}             & 73.51      & \underline{72.95} &	74.16 & 73.23          & 84.06      & 84.22	& 83.63	& 83.03          & 94.97      & \textbf{92.96} & \underline{92.11}	& 91.70                 \\
\cmidrule{2-14}
&DGA-FOA (ours)      & 73.51      & 73.44 & \underline{72.86} & \underline{71.94} & 84.06      & \underline{83.60} & \underline{83.33} & 82.79          & 94.97      & 93.84                & 93.33                & 92.56                \\
& DGA-FDA (ours)      & 73.51      & \textbf{72.86}      & 73.15      & \textbf{71.73}      & 84.06      &\textbf{83.38}               &\textbf{82.46}                &\underline{82.43}                & 94.97      &93.93   &93.11   &91.27   \\ 
\midrule
\multirow{8}{*}{\textbf{DeepWalk~\cite{perozzi2014deepwalk}}} 
&Meta-Self~\cite{zugner2019adversarial}                & 69.79      & 70.23          & 69.36 & 69.54 & 79.11      & 78.90          & 78.48 & 78.19 & 95.51      & \textbf{93.59} & \underline{92.83} & \underline{92.15} \\
&Meta-Train~\cite{zugner2019adversarial}                & 69.79      & 69.67          & 70.22          & 69.41          & 79.11      & 79.14          & 79.54          & 78.71          & 95.51      & 94.37          & 93.71          & 92.94          \\
&A-Meta-Self~\cite{zugner2019adversarial}             & 69.79      & 69.95 & 69.08          & 69.48          & 79.11      & 78.25          & 79.02          & 78.44 & 95.51      & 94.71          & 93.33          & 92.75          \\
&A-Meta-Train~\cite{zugner2019adversarial}             & 69.79      & 69.93          & 69.92          & 70.76          & 79.11      & 79.37          & 79.29          & 78.64          & 95.51      & 94.70          & 94.65          & 93.94          \\
&GraD~\cite{liu2022towards}             & 69.79      & 70.21	& 70.59	& 70.92          & 79.11      & 79.65	& 79.28	& 78.25         & 95.51      & \underline{94.06} &	\textbf{92.53}	& \textbf{92.11}          \\
\cmidrule{2-14}
&DGA-FOA (ours)      & 69.79      & \underline{68.88} & \underline{67.60} & \underline{65.48} & 79.11      & \textbf{77.64} & \underline{77.22} & \underline{75.48} & 95.51      & 94.41          & 93.56          & 92.77          \\
&DGA-FDA (ours)      & 69.79      & \textbf{68.55} & \textbf{66.92} & \textbf{65.22} & 79.11      & \underline{77.82}          & \textbf{76.66} & \textbf{75.45} & 95.51      & 94.55          & 94.14          & 92.37         \\ 
\midrule
\midrule
\multirow{8}{*}{\textbf{GCN-SVD~\citep{entezari2020all}}} 
&DICE~\cite{waniek2018hiding}                  & 67.36      & 66.27          & 66.11 & 65.73          & 77.65      & 72.34                & 71.77                & 71.38                & 93.99      & \underline{93.19}                & \textbf{91.91}                & \underline{91.00}                   \\
&Meta-Self~\cite{zugner2019adversarial}                & 67.36      & 65.76          & 66.08          & 65.88 & 77.65      & 72.24                & 72.25                & 71.80        & 93.99      & \textbf{93.15}       & 93.03       & 92.94       \\
&Meta-Train~\cite{zugner2019adversarial}                & 67.36      & 66.05          & 66.23          & 65.88          & 77.65      & 72.37                & 72.43                & 72.05                & 93.99      & 93.74                & 93.74                & 93.66                \\
&A-Meta-Self~\cite{zugner2019adversarial}             & 67.36      & 66.25 & 66.47          & 66.25          & 77.65      & 72.38                & 72.47                & 72.41       & 93.99      & 93.52                & \underline{92.27}                & \textbf{90.75}                \\
&A-Meta-Train~\cite{zugner2019adversarial}             & 67.36      & 66.44          & 66.22          & 66.08          & 77.65      & 72.17                & 72.07                & 72.17                & 93.99      & 94.11                & 94.13                & 92.89                \\
&GraD~\cite{liu2022towards}             & 67.36      & 66.29	& 66.53	& 66.72          & 77.65      & 72.39	& 72.16	& 72.13                & 93.99      & 93.46	& 93.10	& 92.55                \\
\cmidrule{2-14}
&DGA-FOA (ours)      & 67.36      & \textbf{65.37} & \underline{65.01} & \underline{65.26} & 77.65      & \underline{72.10}        & \textbf{71.42}       & \textbf{71.26}       & 93.99      & 93.56                & 92.88                & 92.68                \\
&DGA-FDA (ours)      & 67.36      & \underline{65.75}      & \textbf{65.98}      & \textbf{65.23}      & 77.65      &\textbf{71.85}   &\underline{71.50}   &\underline{71.35}   & 93.99      &93.51   &93.04   &92.42   \\ 
\midrule
\multirow{8}{*}{\textbf{GCN-Jaccard~\citep{xu2019topology}}} 
&DICE~\cite{waniek2018hiding}                  & 72.29      & 71.63                         & 71.29                & 70.77                         & 82.59      & 82.32          & 81.92          & 81.69 & -- &--   &--   &--          \\
&Meta-Self~\cite{zugner2019adversarial}                & 72.29      & 71.91                         & \textbf{70.52}                & \textbf{68.87}                & 82.59      & 81.88          & \textbf{80.06} & \textbf{78.75} & -- &--   &--   &-- \\
&Meta-Train~\cite{zugner2019adversarial}                & 72.29      & 71.48                         & 71.36                         & 71.27                          & 82.59      & 81.83          & 81.37          & 80.79          & -- &--   &--   &-- \\
&A-Meta-Self~\cite{zugner2019adversarial}             & 72.29      & 71.44                & 71.13                         & 70.84                         & 82.59      & 82.59          & 81.91          & 81.21 & -- &--   &--   &-- \\
&A-Meta-Train~\cite{zugner2019adversarial}             & 72.29      & \underline{71.27}                         & 71.43                         & 71.08                         & 82.59      & 81.97          & 81.32          & \underline{80.46}          & -- &--   &--   &-- \\
&GraD~\cite{liu2022towards}             & 72.29      & 71.60	& 71.33	& 71.34                         & 82.59      & 82.33	& 82.11	& 81.57          & -- &--   &--   &-- \\
\cmidrule{2-10}
&DGA-FOA (ours)      & 72.29      & \textbf{71.24}                & 70.81                & 70.04                & 82.59      & \textbf{81.49} & \underline{81.02} & 80.78 & -- &--   &--   &-- \\
&DGA-FDA (ours)      & 72.29      &71.41   &\underline{70.76}   &\underline{70.01}  & 82.59      & \underline{81.63}                &81.25                &80.56   & -- &--   &--   &--         \\ \bottomrule    
\end{tabular}}
\end{table*}

\subsection{Attack Performance}
\label{sec:attack}

Table~\ref{tab:main-res} presents the test accuracy results of the proposed DGA method compared to baselines. The accuracy values for the original clean graphs are reported under the 0\% perturbation rate column. Remarkably, DGA outperforms or closely approximates the best performance in 8 out of 9 metrics, while exhibiting a significantly reduced training time (over 6 times faster) and GPU memory usage (over 11 times less) compared to the previous SOTA Meta-Self and Meta-Train methods. 
Additionally, DGA demonstrates strong attack performance even at low perturbation rates, causing a mere 3\% to 4\% drop in test accuracy with only a 1\% budget. Notably, the DGA-FOA method consistently achieves better results than its counterpart, DGA-FDA, on 7 out of 9 metrics. This observation suggests that the first-order approximation exhibits greater robustness and efficacy in practical applications. Detailed descriptions and reproduction information for these baselines can be found in Appendix~\ref{appendix:baseline}. 
Note that the GraD paper~\cite{liu2022towards} evaluates their performance on the overall graph, distinct from the commonly used LCC setting.
Standard deviations of these experiments are provided in Appendix~\ref{appendix: results_std}.
\vspace{3pt} \noindent 
\textbf{Generalizability and Transferability.}
In order to be consistent with the gray-box attack setting, in which attackers have no information on the victim model, we evaluate the generalizability and transferability of DGA with two different GNN models. We conduct experiments with both supervised attention-based method GAT~\cite{veličković2018graph} and unsupervised random walk-based method DeepWalk~\cite{perozzi2014deepwalk} trained with the DGA poisoned graph. These graphs are derived using GCN as the surrogate model. Detailed descriptions for these base models are provided in Appendix~\ref{appendix:base_model}.
The results, shown in the GAT and DeepWalk section in Table~\ref{tab:more_res}, demonstrate that our method achieves good transferability performance on both GAT and DeepWalk models.
Our method achieved the best or second-best performance in 11 out of 18 metrics. 
Note that since the graph poisoned by DICE contains isolated nodes, we do not evaluate DeepWalk with this method.

\subsection{Training Time and Memory Usage.}
\label{sec:complex}
Table~\ref{tab: computation_complexity} presents a comparison of training time and maximum GPU memory usage between the proposed DGA method and baselines, considering a 5\% perturbation rate. 
Our approach demonstrates significantly shorter training time and lower memory usage compared to the baselines.
Regarding training time, our DGA methods demonstrate significant efficiency. Specifically, when compared to the previous SOTA methods, DGA achieves training time reductions of over 5, 6, and 16 times on the Citeseer, Cora, and PolBlogs datasets, respectively.
Furthermore, our DGA methods exhibit superior efficiency in terms of GPU memory usage. Compared to the SOTA methods, DGA showcases reductions of over 11, 12, and 15 times on the Citeseer, Cora, and PolBlogs datasets, respectively.
Additional comparisons with other budgets can be found in Appendix~\ref{appendix: computation_complexity}.
It is worth noting that DICE is a random perturbation method that runs quickly and does not utilize GPU resources. 
These findings highlight the computational advantages of DGA, making it a highly efficient and scalable solution for adversarial attacks on graph structures. The substantial reductions in training time and GPU memory usage signify the potential of DGA for practical applications in real-world scenarios, as it can be easily applied to large-scale graphs.

\subsection{Robustness against Existing Defense Methods}
\label{sec:defense}

To evaluate the effectiveness of our proposed DGA attack method, we conduct experiments to test its robustness against existing defense methods. 
By doing so, we aim to assess whether our attack method could overcome these defense methods and reveal vulnerabilities in the system that were previously undiscovered. 
We test DGA against two popular defense methods, namely GCN-Jaccard~\citep{xu2019topology} and GCN-SVD~\citep{entezari2020all}. Detailed descriptions for these defense models are provided in Appendix~\ref{appendix:base_model}.
In these experiments, the poisoned graphs are first vaccinated with defense models. 
Then a GCN model is trained from scratch on the vaccinated graph.
We report the classification accuracy of the well-trained GCN model.
Results are shown in the Low-rank SVD and Jaccard section in Table~\ref{tab:more_res}. 
Note that we cannot perform Jaccard on the PolBlogs dataset, as this dataset does not contain node features.
Our method achieves the best or second-best performance compared with the baseline on 11 out of 15 metrics. These results demonstrate the effectiveness of our DGA attack method in overcoming existing defense methods and exposing potential vulnerabilities of GNN models.

\begin{table}[bp]\centering
\caption{Test accuracy (\%) of the GCN model after training with clean and DGA poisoned graphs optimized with different $k$ values in Gumbel top-$k$ trick on the Cora dataset. The best results are highlighted in \textbf{bold}. }
\label{tab:gumbel_k}
\resizebox{0.45\textwidth}{!}{
\begin{tabular}{lcccc}
\toprule
\textbf{Perturbation Rate (\%)} &\textbf{0} &\textbf{1} &\textbf{3} &\textbf{5} \\\midrule
\textbf{$k=1$} &83.62 &80.06 &78.92 &78.37 \\
\textbf{$k=2$} &83.62 &\textbf{79.89} &\textbf{78.77} &\textbf{78.07} \\
\textbf{$k=3$} &83.62 &79.95 &\textbf{78.77} &78.32 \\
\textbf{$k=4$} &83.62 &80.01 &\textbf{78.77} &78.27 \\
\bottomrule
\vspace{-15pt}
\end{tabular}}
\end{table}

\begin{figure*}[btp]
     \centering
     \subfloat[] 
     {\includegraphics[width=0.33\textwidth]{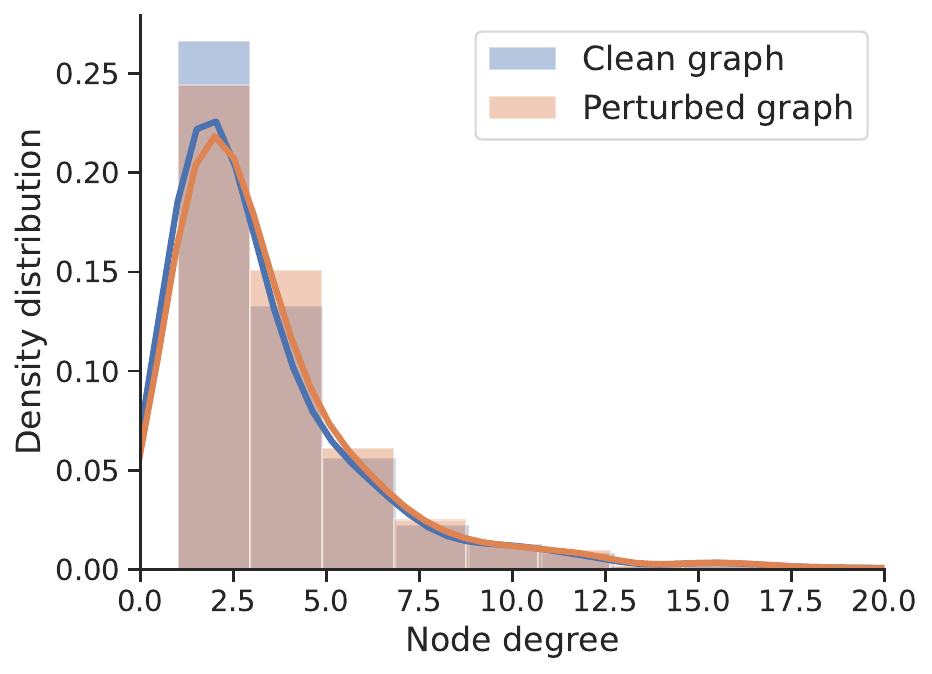}
     \label{fig:degree_dist}}
     \subfloat[] 
     {\includegraphics[width=0.33\textwidth]{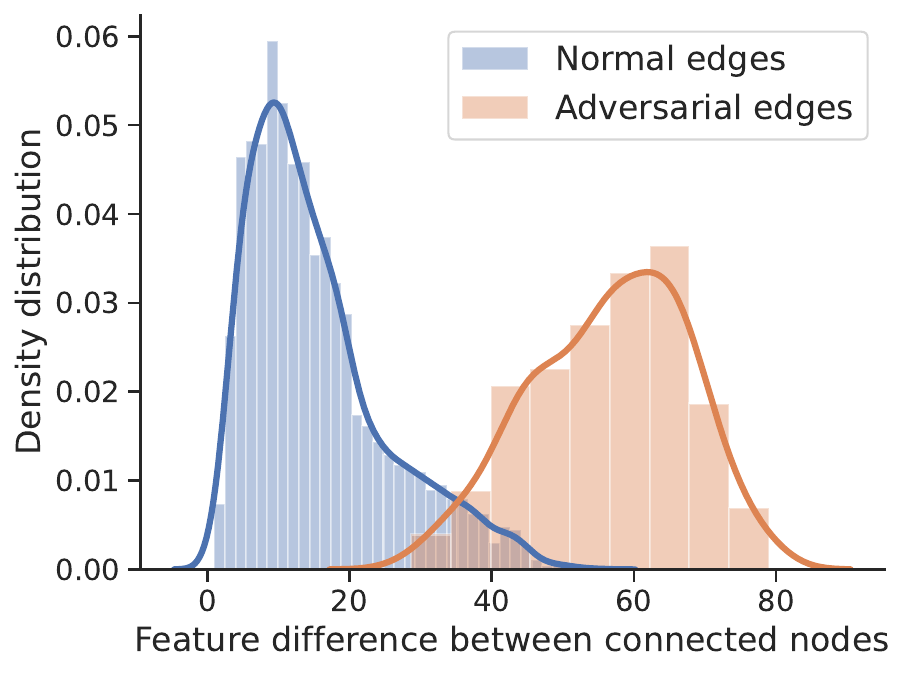}
     \label{fig:feature_diff}}
     \subfloat[] 
     {\includegraphics[width=0.31\textwidth]{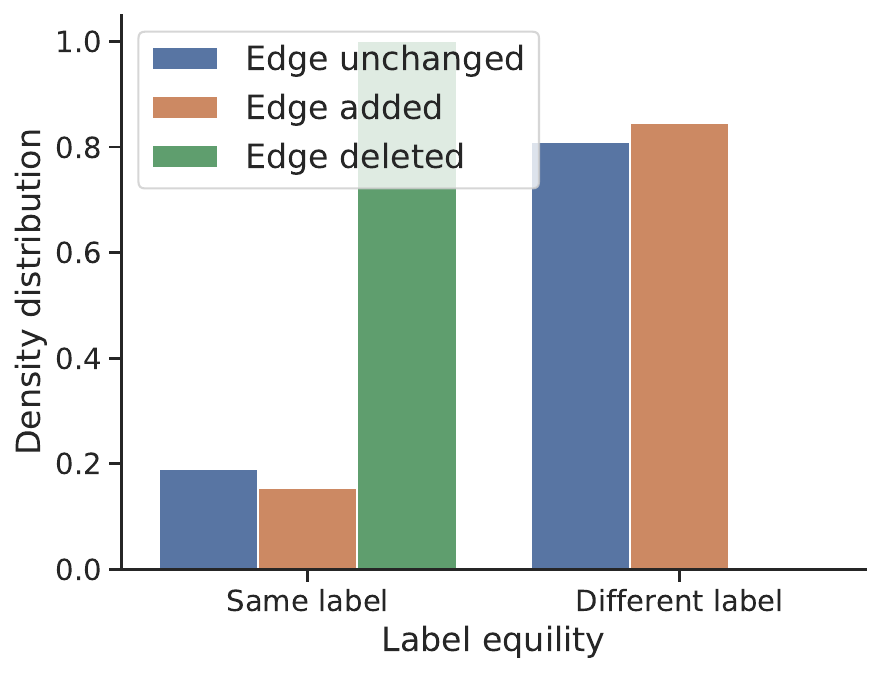}
     \label{fig:label_equal}}
    \caption{ Visualization of statistics of the poisoned graph compared to the original clean graph. Here we provide a comparison on
    \protect\subref{fig:degree_dist} the node degree distribution,
    \protect\subref{fig:feature_diff} the node feature similarity, and
    \protect\subref{fig:label_equal} the label equality between clean and DGA poisoned graphs on the Citeseer dataset. Note that the x-axis of the node degree distribution plot is scaled for better visualization. }
    \label{fig:attack_vis}
\end{figure*} 

\subsection{Impact of Gumbel-\texorpdfstring{$k$}{k}}\label{sec:gumbel-k}
In this section, we conduct ablation studies on the hyperparameter $k$ for the Gumbel-$k$ trick in DGA-FOA using the Cora dataset. As shown in Table~\ref{tab:gumbel_k}, the performance of DGA remains relatively stable across different perturbation rates for varying values of "k." However, we observe a trend where the performance first improves and then declines as $k$ increases. Increasing $k$ initially enhances DGA's performance, indicating its potential for better results. Nevertheless, we find that beyond a certain threshold, the performance starts to decline. This drop in performance can be attributed to overfitting and over-squashing of the GNNs. When $k$ becomes excessively large, the graph becomes dense and closely resembles a fully-connected graph, leading to over-squashing and reduced generalization. Notably, the average degree of the dataset is around 2, as indicated in Table~\ref{tab:datastat}. The ablation study highlights that the best-performing DGA configurations are achieved when setting $k$ around the average degree of the dataset. While higher $k$ values may improve DGA's ability to exploit complex structures, they come at the cost of increased computational resources. Hence, we suggest choosing $k$ near the average degree of the graph to achieve a good balance between performance and computational efficiency.

\subsection{Visualization and Analysis of DGA Attacked Graphs}\label{sec:visual}
\label{sec:visual}
To elucidate the impact of the attack on the graph structure and its implications for the performance and behavior of the targeted model, we visualize three properties for the clean graph and poisoned graph on the Citeseer dataset, as shown in Figure~\ref{fig:attack_vis}. Additional visualizations on other datasets are provided in Appendix~\ref{appendix: visualization}.
Firstly, the node degree distributions for the clean graph and perturbed graph are shown in Figure~\ref{fig:degree_dist}. 
Notably, we observe that the degree distributions of the two graphs exhibit high similarities. 
This observation provides evidence for the imperceptibility of our proposed DGA attack. 
Furthermore, we present the feature similarity and label equality analysis for different types of edges in the graphs, as shown in Figure~\ref{fig:feature_diff} and Figure~\ref{fig:label_equal}, respectively. 
We can observe that DGA exhibits a tendency to establish connections between nodes that possess dissimilar features and have different labels while removing edges that link nodes sharing similar features and belonging to the same label. These observations align with the homophily property in social networks and citation networks and the pattern demonstrated in~\citet{waniek2018hiding}. 


\section{Related Work}


\textbf{Adversarial Attacks on Graphs} aim to interfere with the performance of graph neural networks by introducing subtle perturbations to graph data. 
Following the taxonomy in recent surveys~\cite{jin2021adversarial, sun2022adversarial}, adversarial attacks on GNNs can be classified based on several factors, including 
the level of the task being performed (\eg 
    node-level tasks~\cite{chang2020restricted, entezari2020all, wu2019adversarial, zugner2018adversarial} 
    or graph-level tasks~\cite{dai2018adversarial, tang2020adversarial}), 
the goal of the attack (\eg 
    targeted~\cite{xu2019topology, zugner2018adversarial} or 
    untargeted/global attack~\cite{zugner2019adversarial}), 
the phase of the attack (\eg 
    poisoning attack~\cite{dai2023unnoticeable, liu2019unified, sun2020adversarial} or
    evasion attack~\cite{chang2020restricted, wang2020evasion}), 
the level of knowledge that the attacker has about the model (\eg
    white-box~\cite{wang2019attacking, wu2019adversarial},
    black-box~\cite{dai2018adversarial, ma2020towards, ma2022adversarial}, 
    or gray-box~\cite{bojchevski2019adversarial, sun2020adversarial, zugner2019adversarial}), 
and the perturbation type of the attack (\eg 
    perturbing node features~\cite{ma2020towards, zugner2018adversarial}, 
    graph structure~\cite{liu2022gradients, zugner2019adversarial}, 
    or injecting new nodes~\cite{sun2020adversarial, tao2021adversarial, zou2021tdgia}). 
Different combinations of these scenarios present unique challenges and require different approaches to design effective attacks and defenses. 

In this paper, we focus on a practical problem setting of \textit{node-level, untargeted, gray-box poisoning} attack on \textit{graph structure}~\cite{jin2021adversarial, liu2022gradients, liu2022towards, wu2019adversarial, xu2019topology, zugner2019adversarial}. 
Previous works formulate this setting as a bi-level optimization problem and try to solve it on the discrete graph structure.
MetAttack~\cite{zugner2019adversarial} first employs a meta-learning approach and treat the input data as a hyper-parameter to learn. They adopt a greedy algorithm to obtain the discrete graph structure by selecting one bit to flip at a time, iteratively for each bit in the perturbation budget.
Building on the meta-learning framework, AtkSE~\cite{liu2022gradients} approximates the continuous distribution of hyper-gradients using discrete edge flipping intervals.
GraD~\cite{liu2022towards} introduced a novel attack objective to address the gradient bias.
However, as illustrated in Figure~\ref{fig:compare_complex}, the meta-gradient calculation process of these approaches requires computing the Hessian matrix and accumulating trained-from-scratch inner-loop iteration, resulting in time and resource-intensive operations that are repeated until the perturbation budget is filled.
To achieve effective attacks while minimizing time and resource costs, we propose a novel approach that relaxes the graph structure and learns a continuous edge probability map with attack gradients. 
By simultaneously updating the graph structure and surrogate model parameters, we can efficiently sample optimal perturbation sets from the learned probability map, enabling continuous budget-constrained search.

\vspace{3pt}\noindent
\textbf{Continuous Relaxation on Graph Structure} was first explored in graph structure learning~\cite{zhu2021survey}.
For instance, LDS~\cite{bepler2018learning} incorporates a probabilistic map as a graph generator to model the adjacency matrix of the graph
DGM~\citep{kazi2022differentiable} augments the graph structure using continuous updates to the graph structure with a differentiable graph module and a diffusion module. 
NeuralSparse~\citep{gilmer2017neural} learns $k$-neighbor subgraphs for robust graph representation learning by selecting at most $k$ edges for each nodes.
PTDNet~\cite{luo2021learning} introduces a denoising map drawn from a Bernoulli distribution and learns to drop task-irrelevant edges. 
In contrast to these previous works, our proposed DGA, taking an adversarial learning perspective, is the first to employ continuous relaxation on graph structure and utilize learnable probabilities for attacking GNNs. 
Notably, \citet{xu2019topology} proposed a method for white-box graph attack with continuous relaxation on the perturbation map and use projected gradient descent to update the perturbations. In contrast, DGA adopts stochastic gradient descent to directly update the graph structure, enabling better flexibility, transferability and interpretibility.

\section{Conclusions, Limitations and Social Impact}
This paper proposes a poisoning attack model on graph-structured data. 
We propose a novel approach DGA, which leverages continuous relaxation and parameterization of the graph structure to generate effective and efficient attacks. 
DGA outperforms or approximates the state-of-the-art performance on a variety of benchmark datasets, with significantly less training time and GPU memory occupation compared to existing methods. 
We also provide extensive analysis of the transferability of our approach to other graph models, as well as its robustness against widely-used defense mechanisms.
These can help assess the robustness of existing GNN methods, as well as guide the development of new defense strategies for adversarial attacks.
Additionally, DGA can be expand to large-scale graphs simply with neighbor-sampling training mechanisms, which remains a direction for future exploration.


\section*{Ethical Considerations}
It is worth noting the potential negative social impact of this work. The proposed poisoning attack model, if misused, can significantly impact the robustness and integrity of GNNs. 
This is a reminder of the importance of protecting the privacy of data, including attributes of nodes, training labels, and graph structure, to prevent malicious exploitation and unauthorized attacks.





\bibliographystyle{plainnat} 
\bibliography{my_references}

\begin{thebibliography}{67}
\providecommand{\natexlab}[1]{#1}
\providecommand{\url}[1]{\texttt{#1}}
\expandafter\ifx\csname urlstyle\endcsname\relax
  \providecommand{\doi}[1]{doi: #1}\else
  \providecommand{\doi}{doi: \begingroup \urlstyle{rm}\Url}\fi

\bibitem[Adamic and Glance(2005)]{adamic2005political}
Lada~A Adamic and Natalie Glance.
\newblock The political blogosphere and the 2004 us election: divided they
  blog.
\newblock In \emph{Proceedings of the 3rd international workshop on Link
  discovery}, pages 36--43, 2005.

\bibitem[Bepler and Berger(2019)]{bepler2018learning}
Tristan Bepler and Bonnie Berger.
\newblock Learning protein sequence embeddings using information from
  structure.
\newblock In \emph{International Conference on Learning Representations}, 2019.
\newblock URL \url{https://openreview.net/forum?id=SygLehCqtm}.

\bibitem[Bojchevski and G{\"u}nnemann(2019)]{bojchevski2019adversarial}
Aleksandar Bojchevski and Stephan G{\"u}nnemann.
\newblock Adversarial attacks on node embeddings via graph poisoning.
\newblock In \emph{International Conference on Machine Learning}, pages
  695--704. PMLR, 2019.

\bibitem[Chang et~al.(2020)Chang, Rong, Xu, Huang, Zhang, Cui, Zhu, and
  Huang]{chang2020restricted}
Heng Chang, Yu~Rong, Tingyang Xu, Wenbing Huang, Honglei Zhang, Peng Cui, Wenwu
  Zhu, and Junzhou Huang.
\newblock A restricted black-box adversarial framework towards attacking graph
  embedding models.
\newblock In \emph{Proceedings of the AAAI Conference on Artificial
  Intelligence}, volume~34, pages 3389--3396, 2020.

\bibitem[Chen et~al.(2022)Chen, Li, Shi, Liu, Zhu, and Zhang]{chen2022graph}
Hanxiong Chen, Yunqi Li, Shaoyun Shi, Shuchang Liu, He~Zhu, and Yongfeng Zhang.
\newblock Graph collaborative reasoning.
\newblock In \emph{Proceedings of the Fifteenth ACM International Conference on
  Web Search and Data Mining}, pages 75--84, 2022.

\bibitem[Dai et~al.(2023)Dai, Lin, Zhang, and Wang]{dai2023unnoticeable}
Enyan Dai, Minhua Lin, Xiang Zhang, and Suhang Wang.
\newblock Unnoticeable backdoor attacks on graph neural networks.
\newblock In \emph{Proceedings of the ACM Web Conference 2023}, pages
  2263--2273, 2023.

\bibitem[Dai et~al.(2018)Dai, Li, Tian, Huang, Wang, Zhu, and
  Song]{dai2018adversarial}
Hanjun Dai, Hui Li, Tian Tian, Xin Huang, Lin Wang, Jun Zhu, and Le~Song.
\newblock Adversarial attack on graph structured data.
\newblock In \emph{International conference on machine learning}, pages
  1115--1124. PMLR, 2018.

\bibitem[Deng and Hooi(2021)]{deng2021graph}
Ailin Deng and Bryan Hooi.
\newblock Graph neural network-based anomaly detection in multivariate time
  series.
\newblock In \emph{Proceedings of the AAAI conference on artificial
  intelligence}, volume~35, pages 4027--4035, 2021.

\bibitem[Di~Giovanni et~al.(2023)Di~Giovanni, Giusti, Barbero, Luise, Lio, and
  Bronstein]{di2023over}
Francesco Di~Giovanni, Lorenzo Giusti, Federico Barbero, Giulia Luise, Pietro
  Lio, and Michael Bronstein.
\newblock On over-squashing in message passing neural networks: The impact of
  width, depth, and topology.
\newblock \emph{arXiv preprint arXiv:2302.02941}, 2023.

\bibitem[Entezari et~al.(2020)Entezari, Al-Sayouri, Darvishzadeh, and
  Papalexakis]{entezari2020all}
Negin Entezari, Saba~A Al-Sayouri, Amirali Darvishzadeh, and Evangelos~E
  Papalexakis.
\newblock All you need is low (rank) defending against adversarial attacks on
  graphs.
\newblock In \emph{Proceedings of the 13th International Conference on Web
  Search and Data Mining}, pages 169--177, 2020.

\bibitem[Fan et~al.(2019)Fan, Ma, Li, He, Zhao, Tang, and Yin]{fan2019graph}
Wenqi Fan, Yao Ma, Qing Li, Yuan He, Eric Zhao, Jiliang Tang, and Dawei Yin.
\newblock Graph neural networks for social recommendation.
\newblock In \emph{The world wide web conference}, pages 417--426, 2019.

\bibitem[Fey and Lenssen(2019)]{Fey/Lenssen/2019}
Matthias Fey and Jan~E. Lenssen.
\newblock Fast graph representation learning with {PyTorch Geometric}.
\newblock In \emph{ICLR Workshop on Representation Learning on Graphs and
  Manifolds}, 2019.

\bibitem[Finn et~al.(2017)Finn, Abbeel, and Levine]{finn2017model}
Chelsea Finn, Pieter Abbeel, and Sergey Levine.
\newblock Model-agnostic meta-learning for fast adaptation of deep networks.
\newblock In \emph{International conference on machine learning}, pages
  1126--1135. PMLR, 2017.

\bibitem[Geisler et~al.(2021)Geisler, Schmidt, {\c{S}}irin, Z{\"u}gner,
  Bojchevski, and G{\"u}nnemann]{geisler2021robustness}
Simon Geisler, Tobias Schmidt, Hakan {\c{S}}irin, Daniel Z{\"u}gner, Aleksandar
  Bojchevski, and Stephan G{\"u}nnemann.
\newblock Robustness of graph neural networks at scale.
\newblock \emph{Advances in Neural Information Processing Systems},
  34:\penalty0 7637--7649, 2021.

\bibitem[Gilmer et~al.(2017)Gilmer, Schoenholz, Riley, Vinyals, and
  Dahl]{gilmer2017neural}
Justin Gilmer, Samuel~S Schoenholz, Patrick~F Riley, Oriol Vinyals, and
  George~E Dahl.
\newblock Neural message passing for quantum chemistry.
\newblock In \emph{Proceedings of the 34th International Conference on Machine
  Learning-Volume 70}, pages 1263--1272. JMLR. org, 2017.

\bibitem[Goodfellow et~al.(2015)Goodfellow, Shlens, and
  Szegedy]{goodfellow2015explaining}
Ian Goodfellow, Jonathon Shlens, and Christian Szegedy.
\newblock Explaining and harnessing adversarial examples.
\newblock In \emph{International Conference on Learning Representations}, 2015.
\newblock URL \url{http://arxiv.org/abs/1412.6572}.

\bibitem[Hamilton et~al.(2017)Hamilton, Ying, and
  Leskovec]{hamilton2017inductive}
Will Hamilton, Zhitao Ying, and Jure Leskovec.
\newblock Inductive representation learning on large graphs.
\newblock \emph{Advances in neural information processing systems}, 30, 2017.

\bibitem[Hu et~al.(2022)Hu, Qu, and Work]{hu2022detecting}
Yue Hu, Ao~Qu, and Dan Work.
\newblock Detecting extreme traffic events via a context augmented graph
  autoencoder.
\newblock \emph{ACM Transactions on Intelligent Systems and Technology (TIST)},
  13\penalty0 (6):\penalty0 1--23, 2022.

\bibitem[Ingraham et~al.(2019)Ingraham, Garg, Barzilay, and
  Jaakkola]{ingraham2019generative}
John Ingraham, Vikas Garg, Regina Barzilay, and Tommi Jaakkola.
\newblock Generative models for graph-based protein design.
\newblock \emph{Advances in Neural Information Processing Systems}, 32, 2019.

\bibitem[Jin et~al.(2020)Jin, Ma, Liu, Tang, Wang, and Tang]{jin2020graph}
Wei Jin, Yao Ma, Xiaorui Liu, Xianfeng Tang, Suhang Wang, and Jiliang Tang.
\newblock Graph structure learning for robust graph neural networks.
\newblock In \emph{Proceedings of the 26th ACM SIGKDD international conference
  on knowledge discovery \& data mining}, pages 66--74, 2020.

\bibitem[Jin et~al.(2021)Jin, Li, Xu, Wang, Ji, Aggarwal, and
  Tang]{jin2021adversarial}
Wei Jin, Yaxing Li, Han Xu, Yiqi Wang, Shuiwang Ji, Charu Aggarwal, and Jiliang
  Tang.
\newblock Adversarial attacks and defenses on graphs.
\newblock \emph{ACM SIGKDD Explorations Newsletter}, 22\penalty0 (2):\penalty0
  19--34, 2021.

\bibitem[Kazi et~al.(2022)Kazi, Cosmo, Ahmadi, Navab, and
  Bronstein]{kazi2022differentiable}
Anees Kazi, Luca Cosmo, Seyed-Ahmad Ahmadi, Nassir Navab, and Michael~M
  Bronstein.
\newblock Differentiable graph module (dgm) for graph convolutional networks.
\newblock \emph{IEEE Transactions on Pattern Analysis and Machine
  Intelligence}, 45\penalty0 (2):\penalty0 1606--1617, 2022.

\bibitem[Kingma and Ba(2014)]{kingma2014adam}
Diederik~P Kingma and Jimmy Ba.
\newblock Adam: A method for stochastic optimization.
\newblock \emph{International Conference on Learning Representation}, 2014.

\bibitem[Kipf and Welling(2017)]{kipf2017semisupervised}
Thomas~N. Kipf and Max Welling.
\newblock Semi-supervised classification with graph convolutional networks.
\newblock In \emph{International Conference on Learning Representations}, 2017.
\newblock URL \url{https://openreview.net/forum?id=SJU4ayYgl}.

\bibitem[Kool et~al.(2019)Kool, Van~Hoof, and Welling]{kool2019stochastic}
Wouter Kool, Herke Van~Hoof, and Max Welling.
\newblock Stochastic beams and where to find them: The gumbel-top-k trick for
  sampling sequences without replacement.
\newblock In \emph{International Conference on Machine Learning}, pages
  3499--3508. PMLR, 2019.

\bibitem[Li et~al.(2020)Li, Jin, Xu, and Tang]{li2020deeprobust}
Yaxin Li, Wei Jin, Han Xu, and Jiliang Tang.
\newblock Deeprobust: A pytorch library for adversarial attacks and defenses.
\newblock \emph{arXiv preprint arXiv:2005.06149}, 2020.

\bibitem[Liu et~al.(2019{\natexlab{a}})Liu, Simonyan, and Yang]{liu2018darts}
Hanxiao Liu, Karen Simonyan, and Yiming Yang.
\newblock {DARTS}: Differentiable architecture search.
\newblock In \emph{International Conference on Learning Representations},
  2019{\natexlab{a}}.
\newblock URL \url{https://openreview.net/forum?id=S1eYHoC5FX}.

\bibitem[Liu et~al.(2019{\natexlab{b}})Liu, Si, Zhu, Li, and
  Hsieh]{liu2019unified}
Xuanqing Liu, Si~Si, Xiaojin Zhu, Yang Li, and Cho-Jui Hsieh.
\newblock A unified framework for data poisoning attack to graph-based
  semi-supervised learning.
\newblock In \emph{Proceedings of the 33rd International Conference on Neural
  Information Processing Systems}, pages 9780--9790, 2019{\natexlab{b}}.

\bibitem[Liu et~al.(2022{\natexlab{a}})Liu, Luo, Wu, Li, Liu, and
  Li]{liu2022gradients}
Zihan Liu, Yun Luo, Lirong Wu, Siyuan Li, Zicheng Liu, and Stan~Z Li.
\newblock Are gradients on graph structure reliable in gray-box attacks?
\newblock In \emph{Proceedings of the 31st ACM International Conference on
  Information \& Knowledge Management}, pages 1360--1368, 2022{\natexlab{a}}.

\bibitem[Liu et~al.(2022{\natexlab{b}})Liu, Luo, Wu, Liu, and
  Li]{liu2022towards}
Zihan Liu, Yun Luo, Lirong Wu, Zicheng Liu, and Stan~Z. Li.
\newblock Towards reasonable budget allocation in untargeted graph structure
  attacks via gradient debias.
\newblock In Alice~H. Oh, Alekh Agarwal, Danielle Belgrave, and Kyunghyun Cho,
  editors, \emph{Advances in Neural Information Processing Systems},
  2022{\natexlab{b}}.
\newblock URL \url{https://openreview.net/forum?id=vkGk2HI8oOP}.

\bibitem[Luo et~al.(2021)Luo, Cheng, Yu, Zong, Ni, Chen, and
  Zhang]{luo2021learning}
Dongsheng Luo, Wei Cheng, Wenchao Yu, Bo~Zong, Jingchao Ni, Haifeng Chen, and
  Xiang Zhang.
\newblock Learning to drop: Robust graph neural network via topological
  denoising.
\newblock In \emph{Proceedings of the 14th ACM international conference on web
  search and data mining}, pages 779--787, 2021.

\bibitem[Ma et~al.(2020)Ma, Ding, and Mei]{ma2020towards}
Jiaqi Ma, Shuangrui Ding, and Qiaozhu Mei.
\newblock Towards more practical adversarial attacks on graph neural networks.
\newblock \emph{Advances in neural information processing systems},
  33:\penalty0 4756--4766, 2020.

\bibitem[Ma et~al.(2022)Ma, Deng, and Mei]{ma2022adversarial}
Jiaqi Ma, Junwei Deng, and Qiaozhu Mei.
\newblock Adversarial attack on graph neural networks as an influence
  maximization problem.
\newblock In \emph{Proceedings of the Fifteenth ACM International Conference on
  Web Search and Data Mining}, pages 675--685, 2022.

\bibitem[Madry et~al.(2018)Madry, Makelov, Schmidt, Tsipras, and
  Vladu]{madry2018towards}
Aleksander Madry, Aleksandar Makelov, Ludwig Schmidt, Dimitris Tsipras, and
  Adrian Vladu.
\newblock Towards deep learning models resistant to adversarial attacks.
\newblock In \emph{International Conference on Learning Representations}, 2018.
\newblock URL \url{https://openreview.net/forum?id=rJzIBfZAb}.

\bibitem[McCallum et~al.(2000)McCallum, Nigam, Rennie, and
  Seymore]{mccallum2000automating}
Andrew~Kachites McCallum, Kamal Nigam, Jason Rennie, and Kristie Seymore.
\newblock Automating the construction of internet portals with machine
  learning.
\newblock \emph{Information Retrieval}, 3:\penalty0 127--163, 2000.

\bibitem[Pang et~al.(2022)Pang, Wu, Shen, Zhang, Wei, Xu, Chang, Long, and
  Pei]{pang2022heterogeneous}
Yitong Pang, Lingfei Wu, Qi~Shen, Yiming Zhang, Zhihua Wei, Fangli Xu, Ethan
  Chang, Bo~Long, and Jian Pei.
\newblock Heterogeneous global graph neural networks for personalized
  session-based recommendation.
\newblock In \emph{Proceedings of the fifteenth ACM international conference on
  web search and data mining}, pages 775--783, 2022.

\bibitem[Paszke et~al.(2019)Paszke, Gross, Massa, Lerer, Bradbury, Chanan,
  Killeen, Lin, Gimelshein, Antiga, et~al.]{paszke2019pytorch}
Adam Paszke, Sam Gross, Francisco Massa, Adam Lerer, James Bradbury, Gregory
  Chanan, Trevor Killeen, Zeming Lin, Natalia Gimelshein, Luca Antiga, et~al.
\newblock {PyTorch}: An imperative style, high-performance deep learning
  library.
\newblock \emph{Advances in neural information processing systems}, 32, 2019.

\bibitem[Perozzi et~al.(2014)Perozzi, Al-Rfou, and Skiena]{perozzi2014deepwalk}
Bryan Perozzi, Rami Al-Rfou, and Steven Skiena.
\newblock Deepwalk: Online learning of social representations.
\newblock In \emph{Proceedings of the 20th ACM SIGKDD international conference
  on Knowledge discovery and data mining}, pages 701--710, 2014.

\bibitem[Rozemberczki et~al.(2021)Rozemberczki, Allen, and
  Sarkar]{rozemberczki2021multi}
Benedek Rozemberczki, Carl Allen, and Rik Sarkar.
\newblock Multi-scale attributed node embedding.
\newblock \emph{Journal of Complex Networks}, 9\penalty0 (2):\penalty0 cnab014,
  2021.

\bibitem[Sen et~al.(2008)Sen, Namata, Bilgic, Getoor, Galligher, and
  Eliassi-Rad]{sen2008collective}
Prithviraj Sen, Galileo Namata, Mustafa Bilgic, Lise Getoor, Brian Galligher,
  and Tina Eliassi-Rad.
\newblock Collective classification in network data.
\newblock \emph{AI magazine}, 29\penalty0 (3):\penalty0 93--93, 2008.

\bibitem[St{\"a}rk et~al.(2022)St{\"a}rk, Ganea, Pattanaik, Barzilay, and
  Jaakkola]{stark2022equibind}
Hannes St{\"a}rk, Octavian Ganea, Lagnajit Pattanaik, Regina Barzilay, and
  Tommi Jaakkola.
\newblock {EquiBind}: Geometric deep learning for drug binding structure
  prediction.
\newblock In \emph{International Conference on Machine Learning}, pages
  20503--20521. PMLR, 2022.

\bibitem[Sun et~al.(2022)Sun, Dou, Yang, Zhang, Wang, Philip, He, and
  Li]{sun2022adversarial}
Lichao Sun, Yingtong Dou, Carl Yang, Kai Zhang, Ji~Wang, S~Yu Philip, Lifang
  He, and Bo~Li.
\newblock Adversarial attack and defense on graph data: A survey.
\newblock \emph{IEEE Transactions on Knowledge and Data Engineering}, 2022.

\bibitem[Sun et~al.(2020)Sun, Wang, Tang, Hsieh, and
  Honavar]{sun2020adversarial}
Yiwei Sun, Suhang Wang, Xianfeng Tang, Tsung-Yu Hsieh, and Vasant Honavar.
\newblock Adversarial attacks on graph neural networks via node injections: A
  hierarchical reinforcement learning approach.
\newblock In \emph{Proceedings of The Web Conference 2020}, WWW '20, page
  673–683, New York, NY, USA, 2020. Association for Computing Machinery.
\newblock ISBN 9781450370233.
\newblock \doi{10.1145/3366423.3380149}.
\newblock URL \url{https://doi.org/10.1145/3366423.3380149}.

\bibitem[Tang et~al.(2020{\natexlab{a}})Tang, Ma, Chen, Guo, Wang, Zeng, and
  Zhan]{tang2020adversarial}
Haoteng Tang, Guixiang Ma, Yurong Chen, Lei Guo, Wei Wang, Bo~Zeng, and Liang
  Zhan.
\newblock Adversarial attack on hierarchical graph pooling neural networks.
\newblock \emph{arXiv preprint arXiv:2005.11560}, 2020{\natexlab{a}}.

\bibitem[Tang et~al.(2020{\natexlab{b}})Tang, Li, Sun, Yao, Mitra, and
  Wang]{tang2020transferring}
Xianfeng Tang, Yandong Li, Yiwei Sun, Huaxiu Yao, Prasenjit Mitra, and Suhang
  Wang.
\newblock Transferring robustness for graph neural network against poisoning
  attacks.
\newblock In \emph{Proceedings of the 13th international conference on web
  search and data mining}, pages 600--608, 2020{\natexlab{b}}.

\bibitem[Tang et~al.(2022)Tang, Liu, He, Wang, and Shah]{tang2022friend}
Xianfeng Tang, Yozen Liu, Xinran He, Suhang Wang, and Neil Shah.
\newblock Friend story ranking with edge-contextual local graph convolutions.
\newblock In \emph{Proceedings of the Fifteenth ACM International Conference on
  Web Search and Data Mining}, pages 1007--1015, 2022.

\bibitem[Tao et~al.(2021)Tao, Shen, Cao, Hou, and Cheng]{tao2021adversarial}
Shuchang Tao, Huawei Shen, Qi~Cao, Liang Hou, and Xueqi Cheng.
\newblock Adversarial immunization for certifiable robustness on graphs.
\newblock In \emph{Proceedings of the 14th ACM International Conference on Web
  Search and Data Mining}, pages 698--706, 2021.

\bibitem[Veličković et~al.(2018)Veličković, Cucurull, Casanova, Romero,
  Liò, and Bengio]{veličković2018graph}
Petar Veličković, Guillem Cucurull, Arantxa Casanova, Adriana Romero, Pietro
  Liò, and Yoshua Bengio.
\newblock Graph attention networks.
\newblock In \emph{International Conference on Learning Representations}, 2018.
\newblock URL \url{https://openreview.net/forum?id=rJXMpikCZ}.

\bibitem[Wang and Gong(2019)]{wang2019attacking}
Binghui Wang and Neil~Zhenqiang Gong.
\newblock Attacking graph-based classification via manipulating the graph
  structure.
\newblock In \emph{Proceedings of the 2019 ACM SIGSAC Conference on Computer
  and Communications Security}, pages 2023--2040, 2019.

\bibitem[Wang et~al.(2020)Wang, Zhou, Lin, Zhou, Li, Pang, Fu, Li, and
  Chen]{wang2020evasion}
Binghui Wang, Tianxiang Zhou, Minhua Lin, Pan Zhou, Ang Li, Meng Pang, Cai Fu,
  Hai Li, and Yiran Chen.
\newblock Evasion attacks to graph neural networks via influence function.
\newblock \emph{arXiv preprint arXiv:2009.00203}, 2020.

\bibitem[Wang et~al.(2021{\natexlab{a}})Wang, Jia, Cao, and
  Gong]{wang2021certified}
Binghui Wang, Jinyuan Jia, Xiaoyu Cao, and Neil~Zhenqiang Gong.
\newblock Certified robustness of graph neural networks against adversarial
  structural perturbation.
\newblock In \emph{Proceedings of the 27th ACM SIGKDD Conference on Knowledge
  Discovery \&amp; Data Mining}, KDD '21, page 1645–1653, New York, NY, USA,
  2021{\natexlab{a}}. Association for Computing Machinery.
\newblock ISBN 9781450383325.
\newblock \doi{10.1145/3447548.3467295}.
\newblock URL \url{https://doi.org/10.1145/3447548.3467295}.

\bibitem[Wang et~al.(2021{\natexlab{b}})Wang, Hu, Wang, He, Sheng, Orgun, Cao,
  Ricci, and Philip]{wang2021graph}
Shoujin Wang, Liang Hu, Yan Wang, Xiangnan He, Quan~Z Sheng, Mehmet~A Orgun,
  Longbing Cao, Francesco Ricci, and S~Yu Philip.
\newblock Graph learning based recommender systems: a review.
\newblock In \emph{30th International Joint Conference on Artificial
  Intelligence, IJCAI 2021}, pages 4644--4652. International Joint Conferences
  on Artificial Intelligence, 2021{\natexlab{b}}.

\bibitem[Waniek et~al.(2018)Waniek, Michalak, Wooldridge, and
  Rahwan]{waniek2018hiding}
Marcin Waniek, Tomasz~P Michalak, Michael~J Wooldridge, and Talal Rahwan.
\newblock Hiding individuals and communities in a social network.
\newblock \emph{Nature Human Behaviour}, 2\penalty0 (2):\penalty0 139--147,
  2018.

\bibitem[Wu et~al.(2019)Wu, Wang, Tyshetskiy, Docherty, Lu, and
  Zhu]{wu2019adversarial}
Huijun Wu, Chen Wang, Yuriy Tyshetskiy, Andrew Docherty, Kai Lu, and Liming
  Zhu.
\newblock Adversarial examples for graph data: deep insights into attack and
  defense.
\newblock In \emph{Proceedings of the 28th International Joint Conference on
  Artificial Intelligence}, pages 4816--4823, 2019.

\bibitem[Xie and Ermon(2019)]{xie2019reparameterizable}
Sang~Michael Xie and Stefano Ermon.
\newblock Reparameterizable subset sampling via continuous relaxations.
\newblock In \emph{International Joint Conference on Artificial Intelligence},
  2019.

\bibitem[Xu et~al.(2020)Xu, Li, Jin, and Tang]{xu2020adversarial}
Han Xu, Yaxin Li, Wei Jin, and Jiliang Tang.
\newblock Adversarial attacks and defenses: Frontiers, advances and practice.
\newblock In \emph{Proceedings of the 26th ACM SIGKDD International Conference
  on Knowledge Discovery \& Data Mining}, pages 3541--3542, 2020.

\bibitem[Xu et~al.(2019)Xu, Chen, Liu, Chen, Weng, Hong, and
  Lin]{xu2019topology}
Kaidi Xu, Hongge Chen, Sijia Liu, Pin-Yu Chen, Tsui~Wei Weng, Mingyi Hong, and
  Xue Lin.
\newblock Topology attack and defense for graph neural networks: An
  optimization perspective.
\newblock In \emph{International Joint Conference on Artificial Intelligence}.
  International Joint Conferences on Artificial Intelligence, 2019.

\bibitem[Ying et~al.(2018)Ying, He, Chen, Eksombatchai, Hamilton, and
  Leskovec]{ying2018graph}
Rex Ying, Ruining He, Kaifeng Chen, Pong Eksombatchai, William~L Hamilton, and
  Jure Leskovec.
\newblock Graph convolutional neural networks for web-scale recommender
  systems.
\newblock In \emph{Proceedings of the 24th ACM SIGKDD international conference
  on knowledge discovery \& data mining}, pages 974--983, 2018.

\bibitem[Zhang et~al.(2022)Zhang, Wang, Zhu, Shi, Zhang, and
  Zhou]{zhang2022robust}
Mengmei Zhang, Xiao Wang, Meiqi Zhu, Chuan Shi, Zhiqiang Zhang, and Jun Zhou.
\newblock Robust heterogeneous graph neural networks against adversarial
  attacks.
\newblock In \emph{Proceedings of the AAAI Conference on Artificial
  Intelligence}, volume~36, pages 4363--4370, 2022.

\bibitem[Zhang and Zitnik(2020)]{zhang2020gnnguard}
Xiang Zhang and Marinka Zitnik.
\newblock Gnnguard: Defending graph neural networks against adversarial
  attacks.
\newblock In \emph{Proceedings of Neural Information Processing Systems,
  NeurIPS}, 2020.

\bibitem[Zhao et~al.(2022)Zhao, Zheng, Zhuang, Li, and Zeng]{zhao2022joint}
Kai Zhao, Yukun Zheng, Tao Zhuang, Xiang Li, and Xiaoyi Zeng.
\newblock Joint learning of e-commerce search and recommendation with a unified
  graph neural network.
\newblock In \emph{Proceedings of the Fifteenth ACM International Conference on
  Web Search and Data Mining}, pages 1461--1469, 2022.

\bibitem[Zhao and Akoglu(2020)]{Zhao2020PairNorm:}
Lingxiao Zhao and Leman Akoglu.
\newblock Pairnorm: Tackling oversmoothing in gnns.
\newblock In \emph{International Conference on Learning Representations}, 2020.
\newblock URL \url{https://openreview.net/forum?id=rkecl1rtwB}.

\bibitem[Zhu et~al.(2021{\natexlab{a}})Zhu, Ponomareva, Han, and
  Perozzi]{zhu2021shift}
Qi~Zhu, Natalia Ponomareva, Jiawei Han, and Bryan Perozzi.
\newblock Shift-robust gnns: Overcoming the limitations of localized graph
  training data.
\newblock \emph{Advances in Neural Information Processing Systems},
  34:\penalty0 27965--27977, 2021{\natexlab{a}}.

\bibitem[Zhu et~al.(2021{\natexlab{b}})Zhu, Xu, Zhang, Du, Zhang, Liu, Yang,
  and Wu]{zhu2021survey}
Yanqiao Zhu, Weizhi Xu, Jinghao Zhang, Yuanqi Du, Jieyu Zhang, Qiang Liu, Carl
  Yang, and Shu Wu.
\newblock A survey on graph structure learning: Progress and opportunities.
\newblock \emph{arXiv e-prints}, pages arXiv--2103, 2021{\natexlab{b}}.

\bibitem[Zou et~al.(2021)Zou, Zheng, Dong, Guan, Kharlamov, Lu, and
  Tang]{zou2021tdgia}
Xu~Zou, Qinkai Zheng, Yuxiao Dong, Xinyu Guan, Evgeny Kharlamov, Jialiang Lu,
  and Jie Tang.
\newblock Tdgia: Effective injection attacks on graph neural networks.
\newblock In \emph{Proceedings of the 27th ACM SIGKDD Conference on Knowledge
  Discovery \& Data Mining}, pages 2461--2471, 2021.

\bibitem[Z{\"u}gner et~al.(2018)Z{\"u}gner, Akbarnejad, and
  G{\"u}nnemann]{zugner2018adversarial}
Daniel Z{\"u}gner, Amir Akbarnejad, and Stephan G{\"u}nnemann.
\newblock Adversarial attacks on neural networks for graph data.
\newblock In \emph{Proceedings of the 24th ACM SIGKDD international conference
  on knowledge discovery \& data mining}, pages 2847--2856, 2018.

\bibitem[Zügner and Günnemann(2019)]{zugner2019adversarial}
Daniel Zügner and Stephan Günnemann.
\newblock Adversarial attacks on graph neural networks via meta learning.
\newblock In \emph{International Conference on Learning Representations}, 2019.
\newblock URL \url{https://openreview.net/forum?id=Bylnx209YX}.

\end{thebibliography}


\begin{thebibliography}{}

\end{thebibliography}

\setcounter{page}{1}

\clearpage
\appendix                                    

\section{Proof of Theorem~\ref{thm:convergence}} \label{appendix:proofs}

First, we formally state the assumptions of Theorem~\ref{thm:convergence}. 
\begin{assumption}\label{asm:smooth}
For any $\theta,\theta'\in\mathbb{R}^D,\mathbf{q},\mathbf{q}'\in\mathbb{R}^{N^2}$, there exist $G_1,G_2,L_{1,2},L_{2,1},L_{2,2}, H_{1,2}>0$ such that
\begin{align*}
&\|\mathcal{L}_{atk}(\theta,\mathbf{q}) - \mathcal{L}_{atk}(\theta',\mathbf{q})\|_2\leq G_1\|\theta-\theta'\|_2,\\
& \max\left\{ \|\mathcal{L}_{atk}(\theta,\mathbf{q}) - \mathcal{L}_{atk}(\theta,\mathbf{q}')\|_2, \right. \\
&\qquad \left. \|\mathcal{L}_{train}(\theta,\mathbf{q}) - \mathcal{L}_{train}(\theta,\mathbf{q}')\|_2\right\}\leq G_2\|\mathbf{q}-\mathbf{q}'\|_2,\\
&  \|\nabla_2 \mathcal{L}_{atk}(\theta,\mathbf{q})-\nabla_2 \mathcal{L}_{atk}(\theta',\mathbf{q})\|_2 \leq L_{2,1} \|\theta-\theta'\|_2, \\
& \|\nabla_2 \mathcal{L}_{atk}(\theta,\mathbf{q})-\nabla_2 \mathcal{L}_{atk}(\theta,\mathbf{q}')\|_2 \leq L_{2,2}\|\mathbf{q} - \mathbf{q}'\|_2,\\
&  \|\nabla_1 \mathcal{L}_{train}(\theta,\mathbf{q})-\nabla_1 \mathcal{L}_{train}(\theta,\mathbf{q}')\|_2 \leq L_{1,2}\|\mathbf{q} - \mathbf{q}'\|_2,\\
& \|\nabla^2_{1,2}\mathcal{L}_{train}(\theta, \mathbf{q}) - \nabla^2_{1,2}\mathcal{L}_{train}(\theta, \mathbf{q}')\|_F \leq H_{1,2}\|\mathbf{q} - \mathbf{q}'\|_2,
\end{align*}
where $\|\cdot\|_F$ refers to the Frobenius norm.
\end{assumption}

\begin{lemma}\label{lem:smooth-2}
Under Assumption~\ref{asm:smooth}, there exists $G>0$ and $L>0$ such that $\Phi$ is $G-$Lipschitz continuous and $\nabla \Phi$ is $L$-Lipschitz continuous.
\end{lemma}

\begin{proof}
Note that $\nabla \Phi(\mathbf{q}) = \nabla_2 \mathcal{L}_{atk}(\hat{\theta}, \mathbf{q}) - \alpha \nabla^2_{1,2}\mathcal{L}_{train}(\theta, \mathbf{q}) $ $\nabla_1 \mathcal{L}_{atk}(\hat{\theta},\mathbf{q})$. 

First, we can obtain that 
\begin{align*}
    \|\nabla \Phi(\mathbf{q})\|\leq G_2 + \sqrt{\min(D,N^2)} \alpha L_{1,2} G_1.
\end{align*} 
For any $\mathbf{q},\mathbf{q'}$,  we define that $\hat{\theta}'= \theta - \alpha \nabla_{\theta} \mathcal{L}_{train}(\theta,\mathbf{q}')$ such that 
\begin{align*}
& \|\nabla \Phi(\mathbf{q}) - \nabla \Phi(\mathbf{q}')\|_2 \leq \|\nabla_2 \mathcal{L}_{atk}(\hat{\theta}, \mathbf{q}) - \nabla_2 \mathcal{L}_{atk}(\hat{\theta}', \mathbf{q}')\|_2 \\
& \quad\quad\quad\quad\quad\quad\quad + \alpha \|\nabla^2_{1,2}\mathcal{L}_{train}(\theta, \mathbf{q}) \nabla_1 \mathcal{L}_{atk}(\hat{\theta},\mathbf{q}) \\
& \quad\quad\quad\quad\quad\quad\quad - \nabla^2_{1,2}\mathcal{L}_{train}(\theta, \mathbf{q}') \nabla_1 \mathcal{L}_{atk}(\hat{\theta}',\mathbf{q}')\|_2\\
& \leq  \|\nabla_2 \mathcal{L}_{atk}(\hat{\theta}, \mathbf{q}) - \nabla_2 \mathcal{L}_{atk}(\hat{\theta}', \mathbf{q}')\|_2  \\
& \quad\quad + \alpha \|\nabla^2_{1,2}\mathcal{L}_{train}(\theta, \mathbf{q})\|_F \|\nabla_1 \mathcal{L}_{atk}(\hat{\theta},\mathbf{q}) - \nabla_1 \mathcal{L}_{atk}(\hat{\theta}',\mathbf{q}')\|_2\\
& \quad\quad + \alpha \|\nabla^2_{1,2}\mathcal{L}_{train}(\theta, \mathbf{q}) - \nabla^2_{1,2}\mathcal{L}_{train}(\theta, \mathbf{q}')\|_F \| \nabla_1 \mathcal{L}_{atk}(\hat{\theta}',\mathbf{q}')\|_2\\
& \leq (1+\alpha \sqrt{\min(D,N^2)}L_{1,2})\|\nabla_2 \mathcal{L}_{atk}(\hat{\theta}, \mathbf{q}) \\
& \quad\quad - \nabla_2 \mathcal{L}_{atk}(\hat{\theta}', \mathbf{q}')\|_2 + \alpha G_1 H_{1,2} \|\mathbf{q} - \mathbf{q}'\|_2.
\end{align*}
We further have 
\begin{align*}
& \|\nabla_2 \mathcal{L}_{atk}(\hat{\theta}, \mathbf{q}) - \nabla_2 \mathcal{L}_{atk}(\hat{\theta}', \mathbf{q}')\|_2\\
& \leq \|\nabla_2 \mathcal{L}_{atk}(\hat{\theta}, \mathbf{q}) - \nabla_2 \mathcal{L}_{atk}(\hat{\theta}, \mathbf{q}')\|_2 \\
&\quad\quad + \|\nabla_2 \mathcal{L}_{atk}(\hat{\theta}, \mathbf{q}') - \nabla_2 \mathcal{L}_{atk}(\hat{\theta}', \mathbf{q}')\|_2\\
& \leq L_{2,2} \|\mathbf{q} - \mathbf{q}'\|_2 +  L_{2,1} \|\hat{\theta}-\hat{\theta}'\|_2 \\
& = L_{2,2}  \|\mathbf{q} - \mathbf{q}'\|_2 + \alpha L_{2,1}\|  \nabla_{\theta} \mathcal{L}_{train}(\theta,\mathbf{q}) \\
& \quad\quad - \nabla_{\theta} \mathcal{L}_{train}(\theta,\mathbf{q}')\|_2 \leq (L_{2,2}+  \alpha L_{1,2} L_{2,1}) \|\mathbf{q} - \mathbf{q}'\|_2
\end{align*}

We define $G\coloneqq G_2 + \alpha \sqrt{\min(D,N^2)} \, L_{1,2} \, G_1$ and $L \coloneqq (1 + \allowbreak\alpha \sqrt{\min(D,N^2)} \, L_{1,2}) \allowbreak\,(L_{2,2} + \alpha L_{1,2} L_{2,1}) +  \alpha G_1 H_{1,2}$.
\end{proof}

\begin{proof}[Proof of Theorem~\ref{thm:convergence}]
By the $L$-Lipschitzness of $\nabla \Phi$ as shown Lemma~\ref{lem:smooth-2}, we have
\begin{align}\nonumber
    \mathbb{E}_t\Phi(\mathbf{q}_{t+1}) & \leq \Phi(\mathbf{q}_t) - \eta \nabla \Phi(\mathbf{q}_t) \cdot \mathbb{E}_t \nabla\Phi(\tilde{\mathbf{q}}_t) \\ 
    \nonumber
    & \quad\quad + \eta^2 L \mathbb{E}_t \|\nabla\Phi(\tilde{\mathbf{q}}_t) - \nabla \Phi(\mathbf{q}_t)\|_2^2 + \eta^2 L\| \Phi(\mathbf{q}_t)\|_2^2\\
    \nonumber
    & \leq \Phi(\mathbf{q}_t) - \eta \nabla \Phi(\mathbf{q}_t) \cdot \mathbb{E}_t \nabla\Phi(\tilde{\mathbf{q}}_t)\\ 
    & \quad\quad + \eta^2 L^3 \mathbb{E}_t \|\tilde{\mathbf{q}}_t - \mathbf{q}_t\|_2^2 + \eta^2 L\| \Phi(\mathbf{q}_t)\|_2^2,
    \label{eq:smoothness_starter}
\end{align}
	
where $\mathbb{E}_t[\cdot]$ refers to the expectation conditioned on the randomness before iteration $t$. Young's inequalities leads to
\begin{align}\nonumber
& - \eta \nabla \Phi(\mathbf{q}_t) \cdot \mathbb{E}_t \nabla\Phi(\tilde{\mathbf{q}}_t) \\
\nonumber
& = - \eta \| \nabla \Phi(\mathbf{q}_t)\|_2^2 - \eta \mathbb{E}_t\left[\nabla \Phi(\mathbf{q}_t) \cdot (\nabla\Phi(\tilde{\mathbf{q}}_t)-\nabla \Phi(\mathbf{q}_t))\right]\\
\nonumber
& \leq - \frac{\eta}{2} \| \nabla \Phi(\mathbf{q}_t)\|_2^2 + \frac{\eta}{2} \mathbb{E}_t[\|\nabla\Phi(\tilde{\mathbf{q}}_t)-\nabla \Phi(\mathbf{q}_t))\|_2^2]\\
\label{eq:cross_term}
& \leq - \frac{\eta}{2} \| \nabla \Phi(\mathbf{q}_t)\|_2^2 + \frac{\eta L^2}{2} \mathbb{E}_t[\|\tilde{\mathbf{q}}_t-\mathbf{q}_t\|_2^2].
\end{align}
Plug \eqref{eq:cross_term} into \eqref{eq:smoothness_starter} and re-arrange the terms.
\begin{align*}
& \mathbb{E}_t\Phi(\mathbf{q}_{t+1}) \\
& \leq \Phi(\mathbf{q}_t) -\frac{\eta}{2}\left(1-2\eta L\right)\| \nabla \Phi(\mathbf{q}_t)\|_2^2 + \frac{\eta L^2}{2}\left(1+2\eta L\right) \mathbb{E}_t \|\tilde{\mathbf{q}}_t - \mathbf{q}_t\|_2^2.
\end{align*}
Set $\eta = \frac{1}{4L}$ and use the tower property of expectation.
\begin{align*}
\mathbb{E}\|\nabla \Phi(\mathbf{q}_t)\|_2^2	\leq 16 L\mathbb{E}[\Phi(\mathbf{q}_t) - \Phi(\mathbf{q}_{t+1})] + 3L^2 \mathbb{E} \|\tilde{\mathbf{q}}_t - \mathbf{q}_t\|_2^2. 
\end{align*}
Do telescoping sum from $t=0$ to $T-1$ and divide $T$ on both sides.
\begin{align*}
	\frac{1}{T}\sum_{t=0}^{T-1} \mathbb{E}\|\nabla \Phi(\mathbf{q}_t)\|_2^2 & \leq \frac{16 L(\Phi(\mathbf{q}_0) - \inf_{\mathbf{q}}\Phi )}{T} + \frac{3L^2}{T}\sum_{t=0}^{T-1}\mathbb{E} \|\tilde{\mathbf{q}}_t - \mathbf{q}_t\|_2^2.
\end{align*}
\end{proof}

\section{Experimental Setup}\label{appendix:experiment_setup}
\subsection{Dataset Description} \label{appendix:data}

\textbf{CiteSeer}~\cite{sen2008collective} is a citation network containing 3,312 scientific publications classified into 6 classes. The network consists of 4,732 links, and each publication is represented by a binary word vector indicating the presence or absence of words from a dictionary of 3,703 unique words.

\textbf{Cora}~\cite{mccallum2000automating} is another citation network comprising 2,708 scientific publications classified into seven classes. The network includes 5,429 links, and each publication is represented by a binary word vector indicating the presence or absence of words from a dictionary of 1,433 unique words.

\textbf{PolBlogs}~\cite{adamic2005political} is a graph with 1,490 vertices representing political blogs and 19,025 edges representing links between blogs. The links are automatically extracted from the front pages of blogs. Each vertex is labeled as either liberal or conservative, indicating the political leaning of the blog.

\subsection{Implementation Details.} \label{appendix:implementation}
Our method is implemented using PyTorch~\cite{paszke2019pytorch} and PyTorch Geometric~\cite{Fey/Lenssen/2019} frameworks, with training conducted using the Adam optimizer~\cite{kingma2014adam}. The experiments are carried out on a single NVIDIA RTX A5000 24GB GPU. The search space for model and training hyperparameters can be found in Table~\ref{tab: hyper-para}. Note that the number of finetuning iterations is applicable to DGA-FOA only, whereas for DGA-FDA, it is set to 1, indicating one-step optimization. For all experiments, optimal hyperparameters are selected based on the performance of the validation set.

\begin{table*}[!htbp]\centering
\caption{Model and training hyperparameters for our method on different tasks.}\label{tab: hyper-para}
\resizebox{0.9\textwidth}{!}{
\begin{tabular}{lccc}\toprule
\multirow{2}{*}{\textbf{Hyperparameter}} &\multicolumn{3}{c}{\textbf{Values/Search Space}} \\
\cmidrule{2-4}
&\textbf{CiteSeer} &\textbf{Cora} &\textbf{PolBlogs} \\
\midrule
Momentum & 0.9 & 0.9 & 0.9 \\
\#Iters of attack & 100, 150, 200, 500 & 100, 150, 200 & 100, 150, 200, 500 \\
Attack step size $\eta$ & 1e-2, 5e-3, 1e-3, 5e-4, 1e-4, 5e-5, 1e-5 & 1e-2, 5e-3, 1e-3, 5e-4, 1e-4 & 1e-2, 5e-3, 1e-3, 5e-4, 1e-4 \\
\#Iters of finetuning (DGA-FOA) & 50, 100, 150, 200 & 50, 100, 150, 200 & 50, 100, 150, 200 \\
Finetune step size $\alpha$ & 1e-2, 5e-3, 1e-3, 5e-4, 1e-4, 5e-5, 1e-5 & 1e-2, 5e-3, 1e-3, 5e-4, 1e-4 & 1e-2, 5e-3, 1e-3, 5e-4, 1e-4 \\
Gumbel $\tau$ & 1.0 & 1.0 & 1.0 \\
Gumbel $k$ & 1,2,3,4,5 & 2,3,4,5,6 & 5, 10, 15, 20, 25 \\
\bottomrule
\end{tabular}}
\end{table*}

\subsection{Base Model Description}\label{appendix:base_model}
\textbf{Graph Attention Network (GAT)}~\cite{veličković2018graph} is a neural network architecture that performs graph convolutions using attention mechanisms to selectively aggregate information from neighboring nodes.
It is frequently employed as a foundational layer of defense against adversarial attacks.

\textbf{DeepWalk}~\cite{perozzi2014deepwalk} is an unsupervised learning method that aims to learn low-dimensional representations of nodes in a graph. 
It generates random walks within the graph and applies the skip-gram model to learn node embeddings. 
As DeepWalk is trained in an unsupervised manner without node characteristics or graph convolutions, this transfer setting is more challenging.

\textbf{GCN-Jaccard}~\cite{wu2019adversarial} is a defense method that focuses on identifying and removing adversarial nodes in a graph. 
The Jaccard similarity between the neighborhood sets of two nodes is used to determine if a node is likely to be adversarial. 
By identifying and removing such nodes, GCN-Jaccard aims to improve the robustness of graph models against adversarial attacks.

\textbf{GCN-SVD}~\cite{entezari2020all} is a defense method that mitigates adversarial attacks by approximating the graph Laplacian with a low-rank matrix. By reducing the dimensionality of the graph Laplacian, GCN-SVD aims to preserve the essential structural information while suppressing the influence of potential adversarial perturbations. 

\subsection{Baseline Reproduction Detail}\label{appendix:baseline}
\textbf{DICE}~\cite{waniek2018hiding} (Delete Internally, Connect Externally) is an attack method that focuses on modifying the graph structure to undermine the performance of targeted models. 
It achieves this by randomly connecting nodes with different labels or removing edges between nodes with the same label. 
This manipulation of the graph aims to disrupt the original connectivity patterns and induce misclassification errors in the targeted model. 
For our implementation of the DICE attack, we use the code provided in the  DeepRobust package~\cite{li2020deeprobust}.

\textbf{MetAttack}~\cite{zugner2019adversarial} is an attack method that utilizes meta-learning to solve a bi-level optimization problem. It employs a greedy approach to selectively perturb one edge at a time in order to maximize the adversarial impact on the targeted model. 
The method includes four variants of MetAttack, each employing different loss functions and incorporating first-order approximation: Meta-Train, Meta-Self, A-Meta-Train, and A-Meta-Self. 
Note that the ``Train'' variants use cross-entropy loss on the training set, while the ``Self'' variants use self-training loss with pseudo labels. The ``A-'' variants indicate the use of first-order approximations during the optimization process.
For reproduction, we rely on the code available in the DeepRobust package~\cite{li2020deeprobust} with the default hyperparameters included with the code.

\textbf{GraD}~\cite{liu2022towards} is a recent attack method that leverages the meta-learning framework and introduces a novel attack objective to mitigate gradient bias. 
It claims that it outperforms MetAttack in terms of overall graph performance across the datasets, rather than only focusing on the largest connected component.
For reproduction, we use the official code and the default hyperparameters provided at \url{https://github.com/Zihan-Liu-00/GraD--NeurIPS22}.


\section{Additional Experimental Results} \label{appendix: experiment_result}
\subsection{Results with standard deviation} \label{appendix: results_std}
The standard deviations of the experiments conducted on the GCN, GAT, DeepWalk, and GCN model vaccinated with low-rank SVD approximation (GCN-SVD), and Jaccard (GCN-Jaccard) methods, as described in Section~\ref{sec: exp}, are presented in Tables~\ref{tab:main-res-std} to \ref{tab: defense-svd-std}, respectively.  
The standard deviations are calculated based on 10 runs, providing a reliable estimation of the variation in the experimental results. This indicates that our method consistently performs well and exhibits relatively low variability, thereby highlighting its effectiveness and reliability. 
These results demonstrate the robustness and stability of our proposed method.

\begin{table*}[!htbp]
\centering 
\caption{Standard deviations of test accuracies (\%) for GCN models trained on both clean and poisoned graphs.} 
\label{tab:main-res-std}
\resizebox{0.8\textwidth}{!}{
\begin{tabular}{lcccc|cccc|cccc}  
\toprule
\textbf{Dataset}                & \multicolumn{4}{c|}{\textbf{CiteSeer}}                                  & \multicolumn{4}{c|}{\textbf{Cora}}                             & \multicolumn{4}{c}{\textbf{PolBlogs}}                         \\
\textbf{Perturbation Rate (\%)} & \textbf{0} & \textbf{1}     & \textbf{3}     & \textbf{5}     & \textbf{0} & \textbf{1}     & \textbf{3}     & \textbf{5}     & \textbf{0} & \textbf{1}     & \textbf{3}     & \textbf{5}     \\ 
\midrule
DICE~\cite{waniek2018hiding}                   & 0.6      & 0.67 & 0.60 & 0.81          & 1.16      & 0.98 & 0.99 & 1.13& 0.42         & 0.42 & 0.40 & 0.67          \\
Meta-Self~\cite{zugner2019adversarial}              & 0.6 & 0.87 & 1.30 & 1.91            & 1.16      & 2.12 & 2.28 & 2.29 & 0.42         & 0.53 & 0.98 & 0.41  \\
Meta-Train~\cite{zugner2019adversarial}                & 0.6      & 1.23 & 0.8 & 0.78 & 1.16      & 1.21 & 1.26 & 1.47& 0.42         & 0.53 & 1.53 & 1.38          \\
A-Meta-Self~\cite{zugner2019adversarial}              & 0.6 & 0.85 & 1.20 & 1.29 & 1.16      & 1.08 & 1.37 & 1.25 & 0.42         & 0.66 & 0.7 & 0.83          \\
A-Meta-Train~\cite{zugner2019adversarial}              & 0.6 & 0.68 & 1.06 & 1.31          & 1.16      & 1.08 & 1.22 & 1.25         & 0.42         & 0.59 & 2.66 & 2.72          \\
GraD~\cite{liu2022towards}              & 0.6 & 0.66 & 0.63 & 1.02 & 1.16      & 0.96 & 1.22 & 0.92         & 0.42         & 1.08 & 1.46 & 1.41          \\
\midrule
DGA-FOA (ours)       & 0.6 & 0.49 & 0.56 & 0.61 & 1.16  & 1.22 & 1.1 & 1.1          & 0.42      & 0.6  & 0.88          & 0.91          \\
DGA-FDA (ours)      & 0.6 &0.88 & 0.82 & 0.75 & 1.16      &1.14 & 1.16 & 1.16           & 0.42         & 0.84 & 0.78 & 0.92     \\   \bottomrule 
\end{tabular}}
\end{table*}

\begin{table*}[!htbp] 
\caption{Standard deviations of test accuracies (\%) for GAT models after training with clean and poisoned graphs.  }
\label{tab:transfer-gat-std}
\resizebox{0.8\textwidth}{!}{\begin{tabular}{lcccc|cccc|cccc} \toprule
\textbf{Dataset}                & \multicolumn{4}{c|}{\textbf{CiteSeer}}                                  & \multicolumn{4}{c|}{\textbf{Cora}}                            & \multicolumn{4}{c}{\textbf{PolBlogs}}                                           \\
\textbf{Perturbation Rate (\%)} & \textbf{0} & \textbf{1}     & \textbf{3}     & \textbf{5}     & \textbf{0} & \textbf{1}    & \textbf{3}     & \textbf{5}     & \textbf{0} & \textbf{1}           & \textbf{3}           & \textbf{5}           \\ \midrule
DICE~\cite{waniek2018hiding}                   & 1.00 & 0.53 & 0.96 & 0.94 & 0.84 & 0.76 & 0.99 & 0.9 & 0.51 & 0.62 & 0.73 & 0.49                \\
Meta-Self~\cite{zugner2019adversarial}                & 1.00 & 0.8 & 0.8 & 1.12 & 0.84 & 1.1 & 0.86 & 0.56 & 0.51 & 0.56 & 0.67 & 0.88       \\
Meta-Train~\cite{zugner2019adversarial}                & 1.00 & 1.36 & 0.72 & 1.21 & 0.84 & 0.94 & 0.84 & 1.07 & 0.51 & 0.67 & 1.15 & 1.27                \\
A-Meta-Self~\cite{zugner2019adversarial}             &1.00 & 1.37 & 1.05 & 0.89 & 0.84 & 0.95 & 1.21 & 0.76 & 0.51 & 0.61 & 1.02 & 1.14                 \\
A-Meta-Train~\cite{zugner2019adversarial}             & 1.00 & 0.89 & 1.13 & 1.08 & 0.84 & 0.97 & 0.81 & 0.84 & 0.51 & 0.74 & 0.86 & 0.62                 \\
GraD~\cite{liu2022towards}             & 1.00 & 1.17 & 0.82 & 1.11 & 0.84 & 0.7 & 1.04 & 0.79 & 0.51 & 0.49 & 0.66 & 0.54                \\
\midrule
DGA-FOA (ours)      & 1.00      & 0.4 & 0.85 & 0.94 & 0.84      & 0.87 & 1.07 & 0.64          & 0.51      & 0.82& 0.55 & 0.37                \\
DGA-FDA (ours)      & 1.00      &0.87 & 1.08 & 1.09       & 0.84      & 0.7 & 0.47 & 0.64              & 0.51      &0.53 & 0.58 & 0.65 \\ \bottomrule
\end{tabular}}
\end{table*}

\begin{table*}[!htbp]
\caption{Standard deviations of test accuracies (\%) for DeepWalk models after training with clean and poisoned graphs.} \label{tab:transfer-deepwalk-std}
\resizebox{0.8\textwidth}{!}{\begin{tabular}{lcccc|cccc|cccc} \toprule
\textbf{Dataset}                & \multicolumn{4}{c|}{\textbf{CiteSeer}}                                  & \multicolumn{4}{c|}{\textbf{Cora}}                             & \multicolumn{4}{c}{\textbf{PolBlogs}}                         \\
\textbf{Perturbation Rate (\%)} & \textbf{0} & \textbf{1}     & \textbf{3}     & \textbf{5}     & \textbf{0} & \textbf{1}     & \textbf{3}     & \textbf{5}     & \textbf{0} & \textbf{1}     & \textbf{3}     & \textbf{5}     \\ \midrule
Meta-Self~\cite{zugner2019adversarial}                & 0.95 & 1.03 & 1.57 & 0.9 & 0.97 & 0.62 & 0.79 & 0.7 & 0.42 & 0.5 & 0.52 & 0.93 \\
Meta-Train~\cite{zugner2019adversarial}                & 0.95 & 1.0 & 0.67 & 1.71 & 0.97 & 0.9 & 0.63 & 0.93 & 0.42 & 0.65 & 0.65 & 0.79          \\
A-Meta-Self~\cite{zugner2019adversarial}             & 0.95 & 0.94 & 1.5 & 0.98 & 0.97 & 0.67 & 0.8 & 0.58 & 0.42 & 0.54 & 0.51 & 0.59          \\
A-Meta-Train~\cite{zugner2019adversarial}             & 0.95 & 0.83 & 0.88 & 1.27 & 0.97 & 0.86 & 0.93 & 0.73 & 0.42 & 0.65 & 0.58 & 0.57          \\
GraD~\cite{liu2022towards}             & 0.95 & 0.49 & 0.9 & 1.31 & 0.97 & 0.49 & 0.53 & 0.85 & 0.42 & 0.74 & 0.53 & 0.69          \\
\midrule
DGA-FOA (ours)      & 0.95      & 1.11& 0.55& 0.97 & 0.97      & 0.83& 0.8& 1.07 & 0.42     & 0.29& 0.49& 0.39         \\
DGA-FDA (ours)      & 0.95      &1.24 & 0.91 & 1.11 & 0.97      &0.87 & 1.0 & 1.13& 0.42      & 0.43 & 0.69 & 0.54
\\ \bottomrule
\end{tabular}}
\end{table*}

\begin{table*}[!htbp] 
\centering
\caption{Standard deviations of test accuracies (\%) for the GCN model vaccinated with low-rank SVD approximation. }
\label{tab: defense-jaccard-std}
\resizebox{0.8\textwidth}{!}{\begin{tabular}{lcccc|cccc|cccc} \toprule
\textbf{Dataset}                & \multicolumn{4}{c|}{\textbf{CiteSeer}}                                  & \multicolumn{4}{c|}{\textbf{Cora}}                                               & \multicolumn{4}{c}{\textbf{PolBlogs}}                                           \\
\textbf{Perturbation Rate (\%)} & \textbf{0} & \textbf{1}     & \textbf{3}     & \textbf{5}     & \textbf{0} & \textbf{1}           & \textbf{3}           & \textbf{5}           & \textbf{0} & \textbf{1}           & \textbf{3}           & \textbf{5}           \\ \midrule
DICE~\cite{waniek2018hiding}                  & 1.40 & 1.35 & 1.51 & 1.84 & 1.04 & 0.72 & 0.97 & 0.92 & 0.42 & 0.5 & 0.65 & 0.61\\
Meta-Self~\cite{zugner2019adversarial}                & 1.40 & 1.6 & 1.64 & 1.15 & 1.04 & 0.79 & 0.78 & 0.64 & 0.42 & 0.82 & 0.94 & 0.89 \\
Meta-Train~\cite{zugner2019adversarial}                & 1.40 & 1.49 & 1.64 & 2.09 & 1.04 & 0.57 & 0.92 & 0.6 & 0.42 & 0.62 & 0.58 & 0.44                \\
A-Meta-Self~\cite{zugner2019adversarial}             & 1.40 & 1.31 & 1.01 & 1.34 & 1.04 & 0.61 & 0.68 & 0.81 & 0.42 & 0.59 & 0.73 & 1.04               \\
A-Meta-Train~\cite{zugner2019adversarial}             & 1.40 & 1.41 & 1.18 & 1.75 & 1.04 & 0.72 & 0.77 & 0.83 & 0.42 & 0.53 & 0.57 & 1.09               \\
GraD~\cite{liu2022towards}             & 1.40 & 1.3 & 1.28 & 1.11 & 1.04 & 0.62 & 0.72 & 0.82 & 0.42 & 0.61 & 0.57 & 0.78     \\
\midrule
DGA-FOA (ours)      & 1.40      &1.34 & 1.53 & 1.54 & 1.04      &0.45 & 0.72 & 0.52  & 0.42      & 0.61 & 0.6 & 0.75               \\
DGA-FDA (ours)      & 1.40      &0.91 & 1.53 & 1.52      & 1.04      & 0.66 & 0.54 & 0.73   & 0.42      & 0.53 & 0.44 & 0.62   \\ \bottomrule
\end{tabular}}
\end{table*}

\begin{table*}[!htbp]
\centering
\caption{Test accuracy (\%) of the GCN model vaccinated with Jaccard. }\label{tab: defense-svd-std}
\resizebox{0.6\textwidth}{!}{\begin{tabular}{lcccc|cccc} \toprule
\textbf{Dataset}                & \multicolumn{4}{c|}{\textbf{CiteSeer}}                                                                               & \multicolumn{4}{c}{\textbf{Cora}}                             \\
\textbf{Perturbation Rate (\%)} & \textbf{0} & \textbf{1}                    & \textbf{3}                    & \textbf{5}                    & \textbf{0} & \textbf{1}     & \textbf{3}     & \textbf{5}     \\ \midrule
DICE~\cite{waniek2018hiding}                  & 0.99 & 0.87 & 0.82 & 1.12 & 0.98 & 0.95 & 1.04 & 0.85          \\
Meta-Self~\cite{zugner2019adversarial}                & 0.99 & 0.84 & 1.16 & 1.37 & 0.98 & 1.46 & 1.38 & 1.4 \\
Meta-Train~\cite{zugner2019adversarial}                & 0.99 & 0.93 & 1.17 & 1.81 & 0.98 & 0.73 & 0.87 & 1.11\\
A-Meta-Self~\cite{zugner2019adversarial}             & 0.99 & 1.15 & 1.19 & 1.59 & 0.98 & 1.21 & 0.99 & 0.91 \\
A-Meta-Train~\cite{zugner2019adversarial}             & 0.99 & 0.94 & 1.18 & 1.38 & 0.98 & 0.76 & 0.77 & {0.69} \\
GraD~\cite{liu2022towards}             & 0.99 & 0.97 & 0.86 & 1.15 & 0.98 & 1.15 & 1.01 & 0.96          \\
\midrule
DGA-FOA (ours)      & 0.99      & 0.84 & 0.82 & 0.94                & 0.98      &1.06 & 0.95 & 0.51 \\
DGA-FDA (ours)      & 0.99      &0.73 & 1.24 & 0.94  & 0.98      &1.05 & 0.88 & 1.41           \\ \bottomrule    
\end{tabular}}
\end{table*}

\subsection{Computation Complexity with 1\% and 3\% Perturbation Rates} \label{appendix: computation_complexity}

Table~\ref{tab: computation_complexity-addiitional-1} and Table~\ref{tab: computation_complexity-addiitional-3} provide a comparison of training time and peak GPU memory usage between the proposed DGA method and the baselines, considering perturbation rates of 1\% and 3\% respectively. 
Our approach demonstrates notable advantages over baselines, including significantly reduced training time and lower memory usage. Importantly, our method maintains consistent computational efficiency even as the perturbation rate increases. 
These results highlight the computational benefits of DGA, making it an efficient and scalable solution for adversarial attacks on graph structures. Furthermore, the substantial reductions in training time and GPU memory usage also imply the promising potential of DGA for real-world applications, particularly when dealing with large-scale graphs.

\begin{table*}[!htbp]\centering
\caption{Comparison of training time (in seconds), GPU memory occupancy (in MB) after attack between DGA and baselines with 1\% perturbation rate. For a fair comparison, all experiments are conducted on a single 24GB NVIDIA RTX A5000 GPU. }\label{tab: computation_complexity-addiitional-1}
\resizebox{0.6\textwidth}{!}{
\begin{tabular}{lcc|cc|cc}\toprule
\textbf{Dataset} &\multicolumn{2}{c|}{\textbf{Citeseer}} &\multicolumn{2}{c|}{\textbf{Cora}} &\multicolumn{2}{c}{\textbf{PolBlogs}} \\
\textbf{Method} &\textbf{Time} &\textbf{Mem.} &\textbf{Time} &\textbf{Mem.} &\textbf{Time} &\textbf{Mem.} \\
\midrule
Meta-Self~\cite{zugner2019adversarial} &31 &7,045 &33 &7,377 &46 &5,093 \\
Meta-Train~\cite{zugner2019adversarial} &31 &7,045 &33 &7,377 &46 &5,093 \\
A-Meta-Self~\cite{zugner2019adversarial} &82 &2,641 &134 &3,769 &91 &3,289 \\
A-Meta-Train~\cite{zugner2019adversarial} &82 &2,641 &134 &3,769 &91 &3,289 \\
GraD~\cite{liu2022towards} &32 &10,141 &35 &8,899 &43 &5,803 \\
\midrule
DGA-FOA (ours) &14 &1,327 &15 &1,563 &13 &1,109 \\
DGA-FDA (ours) &17 &1,365 &24 &1,635 &8 &1,129 \\
\bottomrule
\end{tabular}}
\end{table*}

\begin{table*}[!htbp]\centering
\caption{Comparison of training time (in seconds), GPU memory occupancy (in MB) after attack between DGA and baselines with 3\% perturbation rate. For a fair comparison, all experiments are conducted on a single 24GB NVIDIA RTX A5000 GPU. }\label{tab: computation_complexity-addiitional-3}
\resizebox{0.6\textwidth}{!}{
\begin{tabular}{lcc|cc|cc}\toprule
\textbf{Dataset} &\multicolumn{2}{c|}{\textbf{Citeseer}} &\multicolumn{2}{c|}{\textbf{Cora}} &\multicolumn{2}{c}{\textbf{PolBlogs}} \\
\textbf{Method} &\textbf{Time} &\textbf{Mem.} &\textbf{Time} &\textbf{Mem.} &\textbf{Time} &\textbf{Mem.} \\
\midrule
Meta-Self~\cite{zugner2019adversarial} &91 &12,345 &104 &14,721 &151 &11,773 \\
Meta-Train~\cite{zugner2019adversarial} &91 &12,345 &104 &14,721 &151 &11,773 \\
A-Meta-Self~\cite{zugner2019adversarial} &251 &5,621 &412 &8,667 &273 &7,749 \\
A-Meta-Train~\cite{zugner2019adversarial} &251 &5,621 &412 &8,667 &273 &7,749 \\
GraD~\cite{liu2022towards} &96 &15,447 &106 &16,243 &138 &12,483 \\
\midrule
DGA-FOA (ours) &14 &1,327 &15 &1,563 &13 &1,109 \\
DGA-FDA (ours) &17 &1,365 &24 &1,635 &8 &1,129 \\
\bottomrule
\end{tabular}}
\end{table*}

\subsection{Visualization for Cora and Polblogs Datasets} \label{appendix: visualization}
\begin{figure*}[!htbp]
     \centering
     \vspace{-10 pt}
     \subfloat[] 
     {\includegraphics[width=0.33\textwidth]{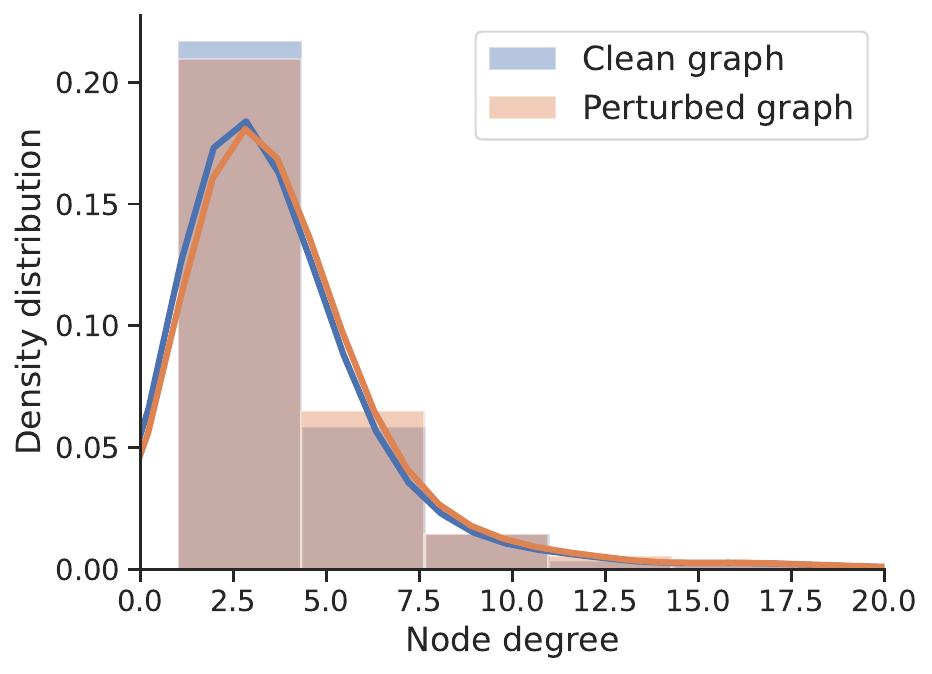}
     \label{fig:degree_dist_cora}}
     \subfloat[] 
     {\includegraphics[width=0.33\textwidth]{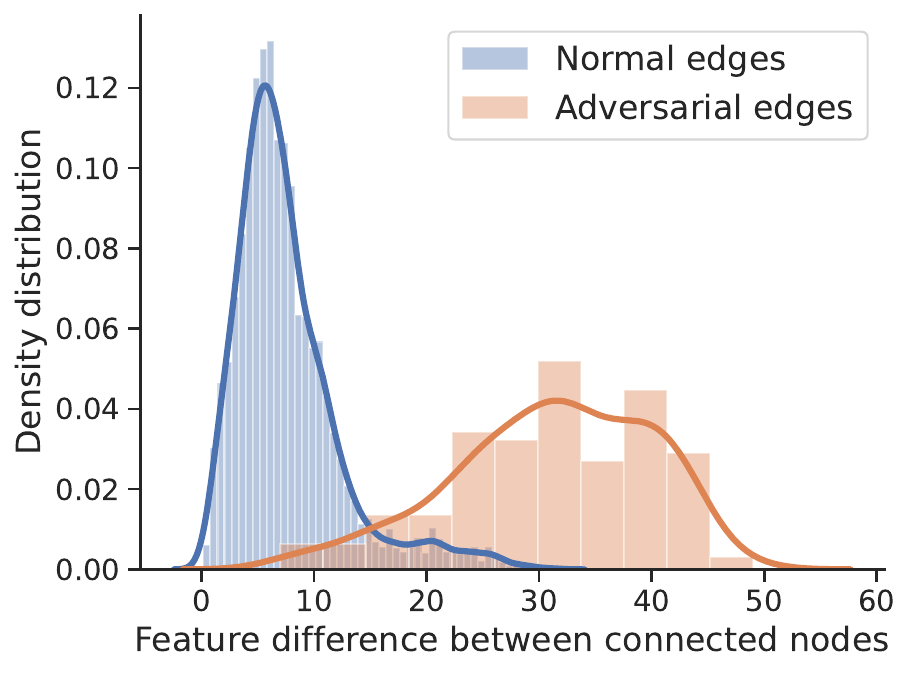}
     \label{fig:feature_diff_cora}}
     \subfloat[] 
     {\includegraphics[width=0.31\textwidth]{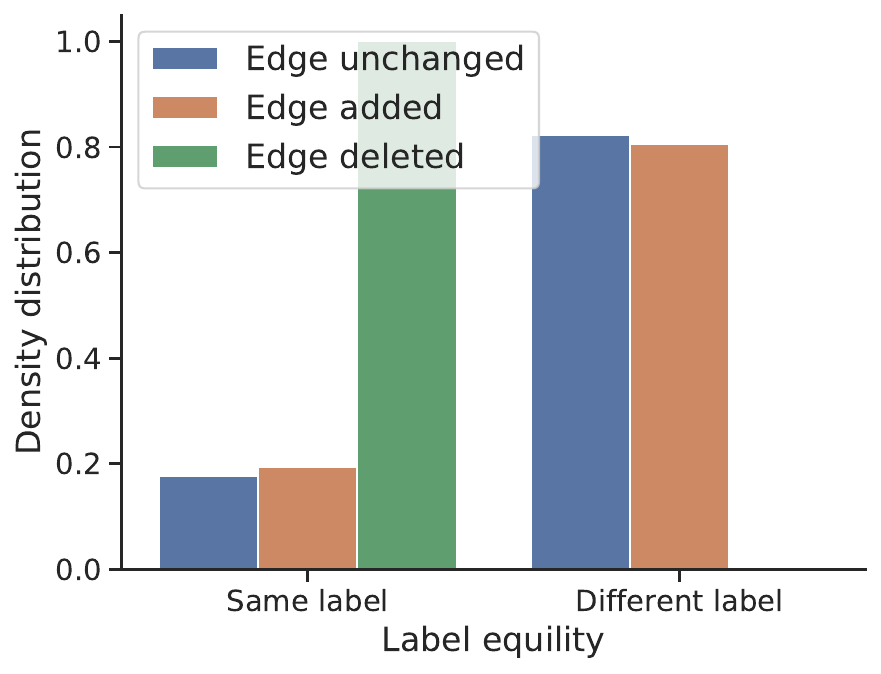}
     \label{fig:label_equal_cora}}
      \vspace{-5 pt}
      \subfloat[] 
     {\includegraphics[width=0.33\textwidth]{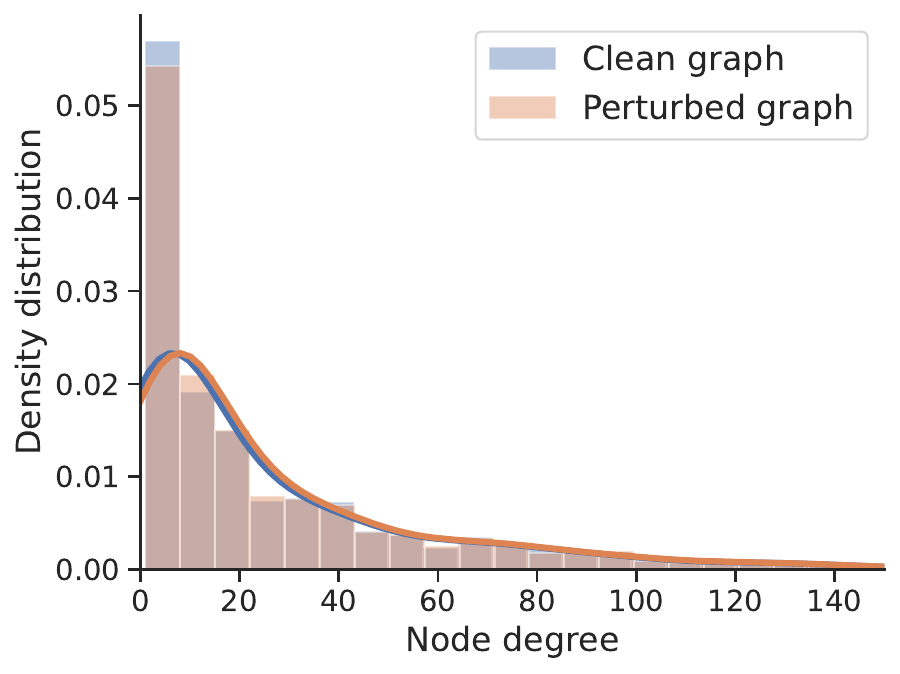}
     \label{fig:degree_dist_polblogs}}
     \subfloat[] 
     {\includegraphics[width=0.33\textwidth]{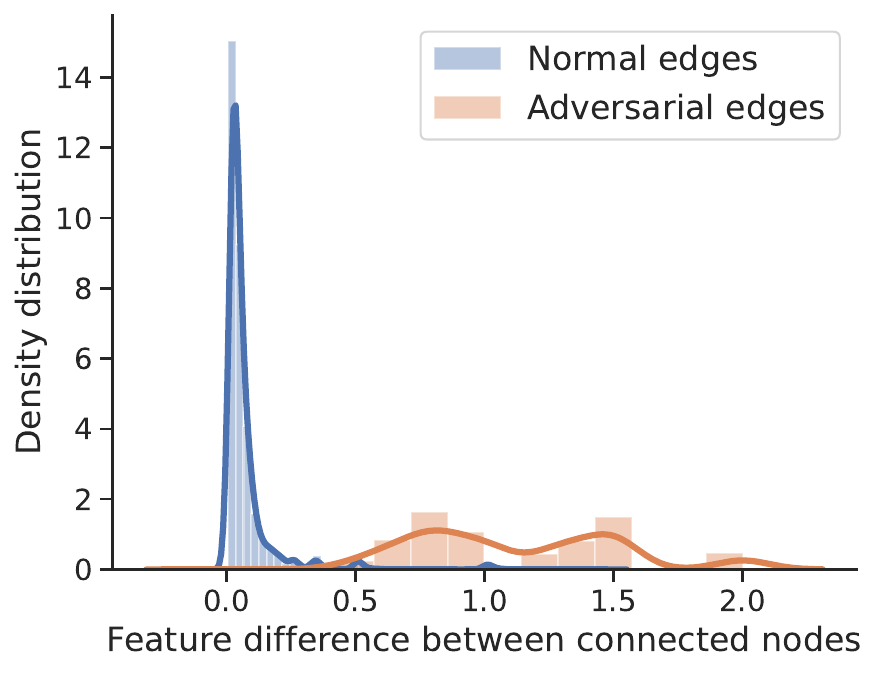}
     \label{fig:feature_diff_polblogs}}
     \subfloat[] 
     {\includegraphics[width=0.31\textwidth]{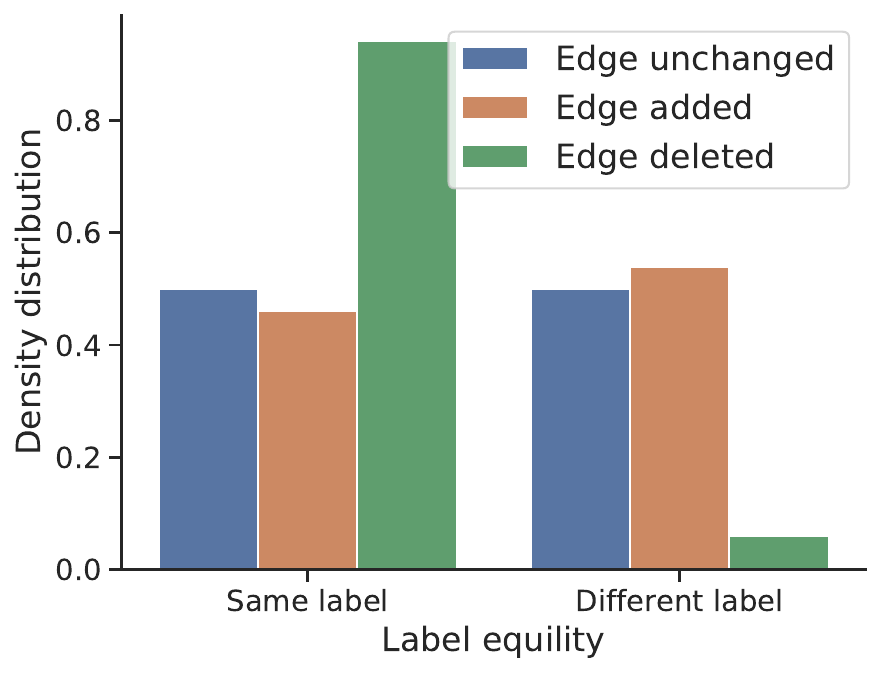}
     \label{fig:label_equal_polblogs}}
      \vspace{-5 pt}
    \caption{ Visualization of statistics of the poisoned graph compared to the original clean graph on the Cora (upper line) and Polblogs (lower line) datasets. Note that the x-axis of the node degree distribution plot is scaled for better visualization. }
    \label{fig:attack_vis_other}
\end{figure*}

We present a visual comparison of statistics between the poisoned graph generated by DGA and the original clean graph. Specifically, we analyze the node degree distribution, node feature similarity, and label equality on the Cora and Polblogs datasets, as shown in Figure~\ref{fig:attack_vis_other}. To enhance clarity and improve visualization, we scale the x-axis of the node degree distribution plot. Notably, we consistently observe similar trends and patterns in these statistics across the Cora and Polblogs datasets, reaffirming our previous observations on the Citeseer dataset as outlined in Sec.~\ref{sec:attack}.
By conducting a visual comparison of these statistics, we gain a better understanding of the attack's influence on the graph structure and its implications for the performance and behavior of the targeted model. This analysis significantly contributes to our comprehensive evaluation and assessment of the effectiveness and impacts of the proposed DGA method.

\end{document}